\documentclass[sigconf,screen,nonacm]{acmart}

\AtBeginDocument{%
  }

\renewcommand\footnotetextcopyrightpermission[1]{}

\usepackage{booktabs}
\usepackage{graphicx}
\usepackage{makecell}
\usepackage{array}
\usepackage{multirow}
\usepackage{tabularx}
\usepackage{adjustbox}
\usepackage{pifont}
\usepackage{threeparttable}
\usepackage{siunitx}
\usepackage{longtable}
\usepackage{seqsplit}

\newcommand{\cmark}{\ding{51}}
\newcommand{\xmark}{\ding{55}}
\begin{document}

\title[Corrective Memory for Drug Discovery Agents]{Constraint-Aware Corrective Memory for Language-Based Drug Discovery Agents}

\author{Maochen Sun}
\affiliation{%
  \institution{Institute of Automation, Chinese Academy of Sciences}
  \country{China}
}
\affiliation{%
  \institution{School of Advanced Interdisciplinary Sciences, University of Chinese Academy of Sciences}
  \country{China}
}
\email{sunmaochen2023@ia.ac.cn}

\author{Youzhi Zhang}
\authornote{Corresponding authors.}
\affiliation{%
  \institution{Centre for Artificial Intelligence and Robotics, Hong Kong Institute of Science \& Innovation, Chinese Academy of Sciences}
  \country{China}
}
\email{youzhi.zhang@cair-cas.org.hk}

\author{Gaofeng Meng}
\authornotemark[1]
\affiliation{%
  \institution{Institute of Automation, Chinese Academy of Sciences}
  \country{China}
}
\affiliation{%
  \institution{Centre for Artificial Intelligence and Robotics, Hong Kong Institute of Science \& Innovation, Chinese Academy of Sciences}
  \country{China}
}
\email{gfmeng@nlpr.ia.ac.cn}


\begin{abstract}
Large language models are making autonomous drug discovery agents increasingly feasible, but reliable success in this setting is not determined by any single action or molecule. It is determined by whether the final returned set jointly satisfies protocol-level requirements such as set size, diversity, binding quality, and developability. This creates a fundamental control problem: the agent plans step by step, while task validity is decided at the level of the whole candidate set. Existing language-based drug discovery systems therefore tend to rely on long raw history and under-specified self-reflection, making failure localization imprecise and planner-facing agent states increasingly noisy. We present \textsc{CACM}  (Constraint-Aware Corrective Memory), a language-based drug discovery framework built around precise set-level diagnosis and a concise memory write-back mechanism. CACM introduces protocol auditing and a grounded diagnostician, which jointly analyze multimodal evidence spanning task requirements, pocket context, and candidate-set evidence to localize protocol violations, generate actionable remediation hints, and bias the next action toward the most relevant correction. To keep planning context compact, \textsc{CACM} organizes memory into static, dynamic, and corrective channels and compresses them before write-back,  thereby preserving persistent task information while  exposing
only the most decision-relevant failures. 
Our experimental results show that CACM improves the target-level success rate by 36.4\% over the state-of-the-art baseline. 
The results show that reliable language-based drug discovery benefits not only from more powerful
molecular tools, but also from more precise diagnosis and more economical agent states.
\end{abstract}



\maketitle
\fancyhf{}
\fancyfoot[C]{\thepage}
\renewcommand{\headrulewidth}{0pt}
\renewcommand{\footrulewidth}{0pt}

\section{Introduction}

Drug discovery is a long, expensive, and highly iterative search process over an enormous chemical space \cite{blass2021drug}. A large body of computational work has therefore sought to accelerate key stages of this pipeline, including structure-based molecular generation, lead optimization, docking, and virtual screening \cite{trott2010autodock,li2021deepligbuilder,peng2022pocket2mol,ragoza2022ligan,jensen2019graphga,tang2024sbddreview}. More recently, deep generative approaches have substantially expanded the design space of structure-based drug discovery, moving from autoregressive and graph-based methods toward diffusion-based 3D molecular generation and pocket-conditioned lead design \cite{guan2023targetdiff,guan2023decompdiff,lin2023d3fg,schneuing2024diffsbdd,huang2024dualdiff,chen2025diffsmol,alakhdar2024diffusionreview}. These advances have made it increasingly realistic to build closed-loop systems that can propose, refine, and evaluate candidate molecules rather than execute only one isolated computational subroutine.

In parallel, large language models (LLMs) have progressed from static text generators to tool-using agents that can plan, invoke external modules, inspect intermediate feedback, and revise subsequent actions \cite{schick2023toolformer,yao2023react,wu2024autogen,wang2024agentsurvey}. This shift has quickly reached the scientific domain. Recent work has explored AI agents for biomedical discovery at large \cite{gao2024biomedical}, autonomous chemistry and experiment design \cite{boiko2023autonomous,ramos2025chemagents}, chemistry tool-use systems \cite{bran2024augmenting,mcnaughton2024cactus}, and domain-specific assistants for molecular or biological workflows \cite{zhou2024autoba,ghafarollahi2024protagents,ishida2025chatchemts,zheng2025llm_drug_review}. As a result, language-based drug discovery is no longer limited to text-only reasoning: an LLM can now function as a planner that coordinates molecular generation, optimization, screening, and evaluation inside an iterative decision loop \cite{averly2025liddia,ock2026agentd}.

However, reliable success in this setting is fundamentally a \emph{set-level} problem. A drug discovery episode does not succeed merely because one step improves docking or because a few molecules appear promising. It succeeds only when the \emph{returned set} jointly satisfies a protocol over size, diversity, binding quality, and developability. In our setting, this protocol is operationalized through metrics such as Quantitative Estimate of Drug-likeness (QED), synthetic accessibility, Lipinski-style drug-likeness constraints, docking-based affinity proxies, novelty, and diversity \cite{bickerton2012qed,ertl2009sas,lipinski2001drug,trott2010autodock}. This means that locally reasonable actions can still fail globally: an agent may over-optimize one property while collapsing diversity, retain too few valid candidates after screening, or return a compact pool that still violates the overall protocol.

This mismatch creates a control problem that current language-based pipelines still handle imprecisely. Existing systems often leave failure interpretation to the planner itself, which must infer from a long accumulated trajectory what went wrong, which requirement is currently violated, and which action should come next. In generic agent settings, self-correction is often expressed as free-form reflection, iterative rewriting, or tool-interactive critique \cite{shinn2023reflexion,madaan2023selfrefine,gou2023critic}. Yet in drug discovery, the corrective signal must be more specific: it should be grounded simultaneously in the task requirements, the protein-pocket context, and the current candidate-set evidence. Otherwise, the planner receives feedback that is semantically plausible but operationally weak.

A second problem lies in inefficient management of agent states. Here, \emph{agent states} denote the planner-facing memory written back across iterations, including task requirements, current pool summaries, action trajectories, screening outcomes, and failure feedback. Throughout the paper, we use \emph{agent states} and \emph{planner-facing memory} interchangeably for these written-back contexts. As the agent proceeds through multiple iterations, this state accumulates heterogeneous information: persistent task inputs, transient pool statistics, action trajectories, screening outcomes, and prior failures. Generic agent research has shown that memory organization is central to long-horizon performance \cite{lewis2020rag,wang2024agentsurvey,zhang2025memorysurvey}, but this issue becomes particularly acute in drug discovery because the agent must reason over a compact context while satisfying a multi-constraint molecular protocol. If static requirements, intermediate pool statistics, and obsolete failure records are all written back in the same way, the planner's context becomes long, noisy, and increasingly difficult to use effectively.

This paper argues that reliable language-based drug discovery requires both \emph{precision} and \emph{parsimony}. Precision means converting set-level failure into an explicit, protocol-grounded diagnosis rather than leaving the planner to infer it implicitly. Parsimony means writing back only the most decision-relevant state, so that the planner receives compact agent states instead of long raw history. To this end, we propose \textsc{CACM}, a framework that reorganizes the closed loop around two components: a \textsc{Protocol Audit} followed by a \textsc{Grounded Diagnoser}, and a structured \textsc{Compress $\&$ Write Back} mechanism over static, dynamic, and corrective memory. The audit checks whether the current candidate set satisfies the task protocol. When it does not, the diagnoser jointly analyzes requirement cues, pocket context, and candidate-set evidence to localize the failure, generate a repair hint, and recommend an action bias for the next step. The resulting signal is then compressed and written back into compact agent states.


Our contributions are threefold. First, we formulate the key bottleneck of language-based drug discovery as a set-level control problem. 
Second, we introduce \textsc{CACM}, which couples \textsc{Protocol Audit} and \textsc{Grounded Diagnoser} with structured static, dynamic, and corrective memory plus compressed write-back mechanism. Third,  our experimental results show that CACM improves the target-level success rate by 36.4\% over the state-of-the-art baseline,
showing that more reliable drug discovery agents depend not only on stronger molecular tools, but also on more precise and more economical control signals.
\section{Related Work}

\subsection{Language-Based Scientific Agents}

Tool-augmented LLM agents have recently shown strong promise across scientific discovery settings \cite{wang2024agentsurvey,gao2024biomedical,ramos2025chemagents}. Existing examples include autonomous chemistry and experiment planning \cite{boiko2023autonomous}, chemistry tool-use systems such as ChemCrow and CACTUS \cite{bran2024augmenting,mcnaughton2024cactus}, multi-omic analysis agents \cite{zhou2024autoba}, protein-design agents \cite{ghafarollahi2024protagents}, de novo molecule-design assistants \cite{ishida2025chatchemts}, drug-repurposing or reasoning-oriented agents \cite{inoue2024drugagent}, and more recent modular drug-discovery agents that span multiple in silico subtasks \cite{ock2026agentd}. These works demonstrate that LLMs can coordinate tools, interpret intermediate results, and support multi-step scientific workflows rather than merely answer one-shot prompts.

That said, these systems are not all solving the same problem. Some focus on automated experimentation or chemistry assistance; some operate in omics or protein-design settings; and some emphasize chat-based molecule-design interfaces or repurposing-oriented search. Our task is narrower and more specific: closed-loop, structure-based \emph{de novo} small-molecule discovery from a protein-pocket input, with success defined by whether the final returned set satisfies a protocol over multiple medicinal-chemistry and docking criteria. Among existing language-based systems, the closest task-aligned reference is LIDDiA, because it exposes a comparable planner--tool loop and provides a public benchmark with compatible inputs, tools, and target-level evaluation. The key difference is that LIDDiA mainly relies on the planner's own trajectory interpretation and goal-checking logic, whereas our method introduces an explicit protocol-aware control layer with deterministic returned-set auditing, failure-localized diagnosis, and a compact planner-facing memory interface. We therefore compare directly against LIDDiA as the strongest task-aligned baseline, while positioning our method as a control-layer improvement to planner-guided drug-discovery agents rather than a narrow modification to one specific planner implementation.

\subsection{AI for Structure-Based Molecular Design and Optimization}

Recent progress in structure-based de novo design has been driven by both classical search methods and modern generative models \cite{tang2024sbddreview}. Earlier lines of work include fragment-based or atom-based pocket construction and search-based optimization, while more recent approaches increasingly rely on neural generators that condition directly on protein-pocket structure \cite{jensen2019graphga,li2021deepligbuilder}. Representative generative SBDD methods include LiGAN for receptor-conditioned 3D generation \cite{ragoza2022ligan}, Pocket2Mol for efficient sampling in protein pockets \cite{peng2022pocket2mol}, and a growing family of diffusion-based approaches such as TargetDiff, DecompDiff, D3FG, DiffSBDD, and related models \cite{guan2023targetdiff,guan2023decompdiff,lin2023d3fg,schneuing2024diffsbdd}. Recent developments have further expanded this line toward dual-purpose generation-and-optimization frameworks, shape-conditioned binding-molecule generation, AlphaFold-informed conditioning, and latent-conditioned structure-based diffusion \cite{huang2024dualdiff,chen2025diffsmol,bernatavicius2024alphafold,le2025latentdiff}. Reviews on diffusion-based molecular design now highlight both their power and their remaining limitations for reliable 3D generation and downstream drug-design use \cite{alakhdar2024diffusionreview}.

These methods have significantly improved the quality of candidate proposal and refinement, but they mainly operate at the level of proposal or local optimization. They do not decide when a system should generate new candidates, when it should optimize an existing pool, when it should screen or merge candidate sets, how to diagnose failure of a returned set against a task protocol, or how to maintain compact agent states across multiple iterations. Our framework is therefore complementary to these molecular tools. Rather than replacing them, we treat them as executors inside a higher-level closed loop that reasons over candidate sets, protocol satisfaction, and corrective control.

\subsection{Reasoning, Reflection, and Memory in LLM Agents}

General LLM-agent research has studied tool use, planning, critique, and multi-agent coordination through frameworks such as Toolformer, ReAct, Reflexion, Self-Refine, CRITIC, AutoGen, and Language Agent Tree Search \cite{schick2023toolformer,yao2023react,shinn2023reflexion,madaan2023selfrefine,gou2023critic,wu2024autogen,zhou2023lats}. These methods establish the broader agentic paradigm in which language models act in iterative loops, inspect feedback, and revise their behavior.

At the same time, recent surveys have highlighted memory as a central design dimension in long-horizon agents, including how information should be stored, retrieved, compressed, and written back \cite{wang2024agentsurvey,zhang2025memorysurvey}. Retrieval-augmented generation and explicit non-parametric memory also provide a broader conceptual basis for grounding language models beyond purely parametric context \cite{lewis2020rag}. However, most generic agent frameworks are task-agnostic and rely on free-form reflection or broad retrieval signals. Drug discovery imposes a more constrained requirement: the agent must satisfy a fixed protocol over an evolving \emph{set} of molecules while operating under limited context. Our method therefore emphasizes two properties that remain underexplored in generic agent work for this setting: \emph{precision}, namely protocol-grounded diagnosis tied to requirements, pocket context, and candidate-set evidence; and \emph{parsimony}, namely structured compression of static, dynamic, and corrective memory before write-back. Rather than adding open-ended reflection alone, we convert verification outcomes into compact control signals that are directly actionable in the next planning step.

\section{Method}

\begin{figure*}[t]
    \centering
    \includegraphics[width=\textwidth, trim=0 6.2cm 0 2.8cm, clip]{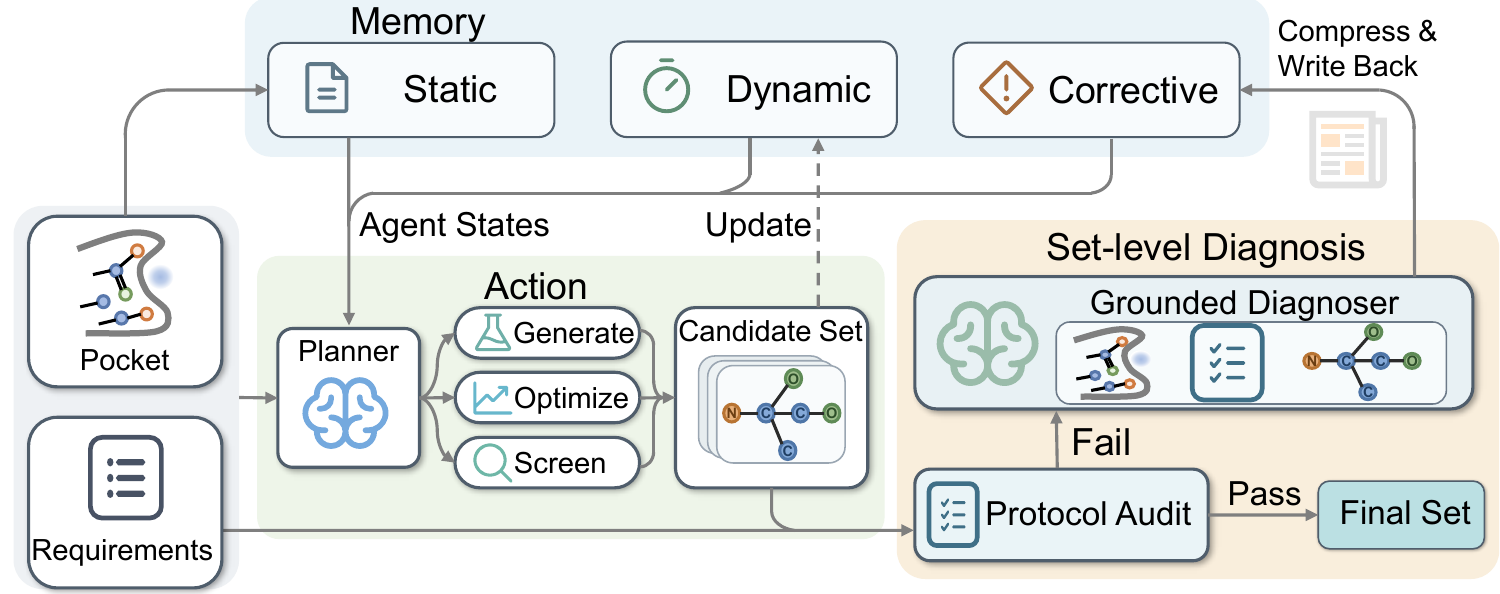}
    \caption{Overview of \textbf{constraint-aware corrective memory (CACM)}.
    CACM maintains an agent state composed of static memory, dynamic descriptive memory, and corrective memory.
    The planner selects one action from \textsc{Generate}, \textsc{Optimize}, and \textsc{Code/Screen}.
    The executed action produces a candidate molecule set, which is checked by a deterministic protocol audit.
    If the current set satisfies the returned-set protocol, it is returned as the final output.
    Otherwise, the audited result is reorganized into corrective memory, compressed, and adapted into the next agent state.
    CACM therefore turns raw history into a compact protocol-aware agent state.}
    \label{fig:method}
\end{figure*}

The central challenge in our setting is not merely to retain more trajectory content, but to expose agent states that are both \emph{precise} and \emph{parsimonious}. The planner must reason over heterogeneous signals, including the returned-set protocol, the receptor-pocket context, summaries of the current candidate pool, and feedback from previous failures. However, these signals are naturally scattered across raw logs, tool outputs, and free-form interaction traces. Directly appending them to memory leads to two coupled problems: the planner receives imprecise decision cues because protocol status and failure causes are not made explicit, and the context length grows rapidly with the number of iterations. As illustrated in Fig.~\ref{fig:method}, CACM addresses these issues through two corresponding components. For \emph{precision}, it makes returned-set failure explicit through deterministic auditing and grounded diagnosis (Sec.~\ref{sec:set_level_diagnosis}). For \emph{parsimony}, it reorganizes heterogeneous write-back signals into functional channels and compresses them into compact agent states for the next decision (Sec.~\ref{sec:concise_memory}).

\subsection{Problem Setup}
\label{sec:problem_setup}

We follow the returned-set design task of LIDDiA~\cite{averly2025liddia}. For a target protein, the input consists of a receptor pocket $P$ and a natural-language requirement set describing the desired properties of the final returned molecule pool. We denote this requirement set by
\begin{equation}
R=\{r_j\}_{j=1}^{m},
\label{eq:requirement_set}
\end{equation}
where each requirement is normalized into a structured tuple
\begin{equation}
r_j=(f_j,\alpha_j,\bowtie_j,b_j).
\label{eq:req_tuple}
\end{equation}
Here $f_j$ denotes the target metric, $\alpha_j$ denotes the corresponding set-level aggregation operator, $\bowtie_j$ is the comparison operator, and $b_j$ is the threshold. Typical examples include cardinality for pool size, a pool-level diversity functional for diversity, and worst-case min/max aggregation for molecule-level properties such as QED, SAScore, Lipinski, novelty, and docking score.

At iteration $k$, the planner chooses one action
\begin{equation}
a^{(k)}=(\tau^{(k)},u^{(k)}),
\label{eq:current_action}
\end{equation}
where $\tau^{(k)} \in \{\textsc{Generate}, \textsc{Optimize}, \textsc{Code/Screen}\}$ is the action type and $u^{(k)}$ is its input. Executing $a^{(k)}$ yields a candidate molecule set $S^{(k)}$ together with pool statistics and molecule-level property summaries for control.

The task is defined at the returned-set level: a run is successful only when the final returned pool jointly satisfies the protocol, rather than merely containing a few individually strong molecules. We therefore define the set-level observation for requirement $r_j$ as
\begin{equation}
o_j\!\left(S^{(k)}\right)=\alpha_j\!\left(f_j,S^{(k)}\right),
\label{eq:set_observation}
\end{equation}
which induces a deterministic acceptance gate
\begin{equation}
g\!\left(S^{(k)},R\right)=1
\iff
\forall j,\; o_j\!\left(S^{(k)}\right)\bowtie_j b_j.
\label{eq:goal_check}
\end{equation}
Otherwise, $g\!\left(S^{(k)},R\right)=0$. That is, the gate returns $1$ exactly when all requirement-specific observations satisfy their corresponding thresholds; otherwise the returned set is rejected.

Rather than planning directly from raw history, CACM conditions the planner on an adapted agent state $\widetilde{M}^{(k)}$:
\begin{equation}
a^{(k+1)}=\mathcal{P}\!\left(\widetilde{M}^{(k)}\right),
\label{eq:planner_with_memory}
\end{equation}
where $\mathcal{P}(\cdot)$ denotes the planner. Equation~\eqref{eq:current_action} defines the action taken at the current iteration, whereas Eq.~\eqref{eq:planner_with_memory} defines how the next action is chosen after the current result has been evaluated and summarized into the next agent state.

The workflow of CACM is as follows: Given a target pocket and a requirement set, the planner first selects one action from \textsc{Generate}, \textsc{Optimize}, and \textsc{Code/Screen}. The molecular toolchain executes this action and produces a candidate molecule set. CACM then checks whether the current set already satisfies the returned-set protocol. If it does, the run terminates and returns the current set. Otherwise, CACM converts the failure into an explicit diagnosis signal, writes this signal back together with the task and search context, and exposes the resulting compact agent state to the planner for the next decision. The role of CACM is therefore not to replace the molecular tools, but to control how protocol-relevant information is diagnosed, organized, and written back across iterations.

\subsection{Set-Level Diagnosis}
\label{sec:set_level_diagnosis}

This subsection addresses the \emph{precision} side of CACM by making returned-set failure explicit through deterministic auditing and grounded diagnosis under a fixed protocol.

CACM first applies a deterministic protocol audit to the current pool. This component is entirely code-based and serves as the source of truth for hard pass/fail decisions at the returned-set level. For each requirement $r_j$, we compute a signed residual
\begin{equation}
\Delta_j^{(k)}=
\begin{cases}
o_j\!\left(S^{(k)}\right)-b_j, & \text{if } \bowtie_j \in \{\ge,>\},\\
b_j-o_j\!\left(S^{(k)}\right), & \text{if } \bowtie_j \in \{\le,<\},
\end{cases}
\label{eq:margin}
\end{equation}
and collect these values into
\begin{equation}
\Delta^{(k)}=\left(\Delta_1^{(k)},\ldots,\Delta_m^{(k)}\right).
\label{eq:margin_vector}
\end{equation}
Under this convention, negative residuals correspond to violated constraints, while their magnitudes quantify the distance to protocol satisfaction for the current pool.

This audit has two roles. First, it determines whether the current pool already satisfies Eq.~\eqref{eq:goal_check}. Second, when the pool still fails, it provides a structured failure description that can be used for corrective write-back. CACM therefore does not rely on an unconstrained language judgment to decide whether a returned set is valid.

CACM then converts the audited failure state into an explicit corrective record through a grounded diagnoser:
\begin{equation}
z^{(k)} = D\!\left(P,R,S^{(k)},\Delta^{(k)}\right),
\label{eq:grounded_diagnoser}
\end{equation}
where $D(\cdot)$ denotes the grounded diagnosis operator. Its inputs are the pocket $P$, the requirement set $R$, the current candidate set $S^{(k)}$, and the residual vector $\Delta^{(k)}$. The output $z^{(k)}$ is a compact corrective record for write-back.

Importantly, $D(\cdot)$ is not an abstract black-box reflection operator. It is grounded by the deterministic audit and by the current set evidence. Given the violated requirements and their residuals, it identifies which requirements remain unsatisfied, determines the dominant failure type, and produces a short repair hint together with a recommended next-action bias. In this way, CACM turns returned-set failure into an explicit and actionable diagnosis signal, rather than leaving the planner to infer the dominant problem from long raw history alone.

\subsection{Concise Memory Construction}
\label{sec:concise_memory}

This subsection addresses the \emph{parsimony} side of CACM by organizing, compressing, and writing back only the most decision-relevant information at each iteration.

A binary fail signal is not sufficient for the next decision. The planner also needs to know what the task requires, what the current search state looks like, and why the previous step did not yet satisfy the protocol. CACM therefore organizes write-back information through three memory channels:
\begin{equation}
M^{(k)}=\big(M_s,\;M_d^{(k)},\;M_c^{(k)}\big),
\label{eq:memory_factorization}
\end{equation}
where $M_s$ is static memory, $M_d^{(k)}$ is dynamic descriptive memory, and $M_c^{(k)}$ is corrective memory.

\paragraph{Static memory.}
Static memory stores target-specific information that remains invariant throughout the trajectory:
\begin{equation}
M_s=\Phi_s(P,R),
\label{eq:static_memory}
\end{equation}
where $\Phi_s(\cdot)$ denotes the static-memory constructor. Given the pocket $P$ and the requirement set $R$, it builds a deterministic target description including the target identity, the requirement set, the pocket file, and a compact pocket summary extracted from the receptor structure. In our system, this pocket summary is computed from lightweight structural descriptors, including pocket size, residue composition, geometric extent, and coarse physicochemical ratios. Static memory therefore provides a stable description of what must ultimately be achieved and what receptor environment is being optimized against.

\paragraph{Dynamic descriptive memory.}
Dynamic descriptive memory stores a compact view of the evolving search state:
\begin{equation}
M_d^{(k)}=\mathcal{T}\!\left(M_d^{(k-1)},\widehat{S}^{(k)},H_a^{(k)}\right),
\label{eq:dynamic_memory}
\end{equation}
where $\mathcal{T}(\cdot)$ denotes the dynamic-memory update operator, $\widehat{S}^{(k)}$ denotes the retained summary of the current pool, and $H_a^{(k)}$ denotes a short recent-action window. Rather than appending the full raw history, CACM keeps only informative pool summaries and a short action history. Concretely, $\widehat{S}^{(k)}$ is constructed by combining a recency cue with a deterministic quality heuristic so that the retained pool remains informative for control, rather than merely logging all previous candidates. $H_a^{(k)}$ records only the most recent decisions and their outcomes, making the current search frontier readable without exposing the planner to the full raw history.

\paragraph{Corrective memory.}
Corrective memory stores failure-localized control information and is updated only when the current pool fails Eq.~\eqref{eq:goal_check} at that step:
\begin{equation}
M_c^{(k)}=\mathcal{U}\!\left(M_c^{(k-1)},z^{(k)}\right),
\label{eq:corrective_memory}
\end{equation}
where $\mathcal{U}(\cdot)$ denotes the corrective-memory update operator, and $z^{(k)}$ is the corrective record produced by the grounded diagnoser in Eq.~\eqref{eq:grounded_diagnoser}. This channel stores the explicitly diagnosed reason for failure and the corresponding repair direction, rather than a free-form history of all past failures.

Explicit memory organization alone is not sufficient. If all historical pool summaries and all diagnostic records are retained indefinitely, the planner input quickly becomes redundant, unstable, and difficult to control. CACM therefore introduces a selection--compression--adaptation pipeline that turns the three memory channels into a compact agent state.

For each channel $x\in\{s,d,c\}$, CACM first applies channel-wise selection before formatting:
\begin{equation}
\bar{M}_x^{(k)}=\operatorname{Sel}_x\!\left(M_x^{(k)}\right),
\qquad x\in\{s,d,c\},
\label{eq:channelwise_selection}
\end{equation}
where $\operatorname{Sel}_x(\cdot)$ denotes the channel-specific selection operator. For the static channel, selection is trivial because the target-level context remains fixed across iterations. For the dynamic channel, $\operatorname{Sel}_d(\cdot)$ retains the most informative pool summaries together with a short recent-action window. For the corrective channel, $\operatorname{Sel}_c(\cdot)$ keeps high-value corrective records rather than the full diagnosis history.

The retained channel contents are then compressed separately:
\begin{equation}
\tilde{M}_x^{(k)}=\mathcal{C}_x\!\left(\bar{M}_x^{(k)}\right),
\qquad x\in\{s,d,c\},
\label{eq:channelwise_compression}
\end{equation}
where $\mathcal{C}_x(\cdot)$ denotes a channel-specific compression operator. Here $\mathcal{C}_x(\cdot)$ is a deterministic formatter rather than a learned summarizer: each channel is rendered into a fixed template. Static compression summarizes the target and pocket context; dynamic compression summarizes the retained pool states and recent actions; corrective compression summarizes the retained failure records together with the repair direction and next-action bias.

Finally, the compressed channels are adapted into the agent state
\begin{equation}
\widetilde{M}^{(k)}
=
\mathcal{A}\!\left(
\tilde{M}_s^{(k)},
\tilde{M}_d^{(k)},
\tilde{M}_c^{(k)}
\right),
\label{eq:memory_adaptation}
\end{equation}
where $\mathcal{A}(\cdot)$ denotes the adaptation operator. The operator $\mathcal{A}(\cdot)$ assembles the labeled channel summaries into a unified agent state. The resulting state therefore remains compact and predictable across iterations, while still preserving the task context, the current search frontier, and the most relevant corrective signal for the next decision step.

\subsection{Protocol-Aware Closed-Loop Control}
\label{sec:closed_loop_control}

CACM therefore operates as a protocol-aware closed loop. At iteration $k$, the planner reads $\widetilde{M}^{(k-1)}$ and selects the next action via Eq.~\eqref{eq:planner_with_memory}. The molecular toolchain executes this action and produces a candidate set $S^{(k)}$. The deterministic protocol audit then evaluates the returned-set status through Eq.~\eqref{eq:goal_check} and computes the residual vector $\Delta^{(k)}$. If the current pool satisfies the protocol, the run terminates under the configured stopping logic. Otherwise, the grounded diagnoser converts the audited failure into a corrective record, the memory channels are updated and compressed, and the next planning step proceeds from the resulting compact agent state.

Within this loop, the planner and the grounded diagnoser have distinct roles. The planner determines which action to execute next, while the grounded diagnoser determines why the current pool still fails and which repair direction should be prioritized. CACM thus couples deterministic protocol checking, failure-localized diagnosis, and concise memory write-back into a unified control loop for returned-set optimization across iterations.


\section{Experiments}

\subsection{Experimental Setup}
\label{sec:exp_setup}

\paragraph{Benchmark and agent setting.}
We evaluate on the same 30-target benchmark introduced by LIDDiA~\cite{averly2025liddia}. Each target provides a curated protein pocket, a natural-language design requirement, and known drugs used to derive target-specific reference thresholds. For all agentic methods, we keep the original action space, namely \textsc{Generate}, \textsc{Optimize}, and \textsc{Code/Screen}, with the same maximum of 10 iterations per target.
Unless otherwise specified, the DeepSeek-based LIDDiA baseline and all CACM variants use \textbf{deepseek-reasoner} as the controller model. The molecular toolchain is unchanged: Pocket2Mol is used for structure-based generation~\cite{peng2022pocket2mol}, GraphGA for molecular optimization~\cite{jensen2019graphga}, and AutoDock Vina for docking-based binding evaluation~\cite{trott2010autodock}.

\paragraph{Baselines.}
Following LIDDiA~\cite{averly2025liddia}, we compare against two task-specific molecular design baselines and four general-purpose LLM baselines. Pocket2Mol is a structure-based generator that uses the target pocket as input, whereas DiffSMOL is a ligand-based generative model that requires a binding ligand. For general-purpose LLMs, we include Claude, GPT-4o, o1-mini, and o1. In Table~\ref{tab:main_results}, the first six columns are taken directly from the original LIDDiA paper, while the last two columns report our DeepSeek-based LIDDiA reproduction and CACM under the same benchmark and molecular tool setting.

\paragraph{Evaluation metrics.}
Following LIDDiA, we report molecule-level quality using drug-likeness (QED)~\cite{bickerton2012qed}, Lipinski compliance (LRF)~\cite{lipinski2001drug}, synthetic accessibility (SAS)~\cite{ertl2009sas}, docking score (VNA)~\cite{trott2010autodock}, novelty (NVT), and set diversity (DVS). A molecule is counted as high-quality (HQ) for a target if it satisfies the same target-specific quality thresholds used in LIDDiA, i.e., it is at least as good as the corresponding known-drug reference on QED, LRF, and docking, no worse on SAS, and also satisfies $\mathrm{NVT}\ge 0.8$.

At the target level, we follow the returned-set evaluation protocol of LIDDiA. Specifically, we report the same target-level indicators as in LIDDiA: $\mathrm{DVS}>0.8$, $N{\geq}5$ \& DVS, $N{\geq}5$ \& HQ, DVS \& HQ, and TSR (\emph{target success rate}). Here, TSR denotes the percentage of targets whose returned sets satisfy the LIDDiA returned-set success conditions. All target-level indicators are computed using the same definitions and target-specific thresholds as in LIDDiA.

\subsection{Main Results}
\label{sec:main_results}

\paragraph{Returned-set performance.}
Table~\ref{tab:main_results} reports the main comparison on the 30-target benchmark. Among the first six baselines, Pocket2Mol and DiffSMOL both achieve perfect diversity coverage at the target level, with $100.0\%$ on DVS $> 0.8$ and on $N{\geq}5$ \& DVS, but their final returned sets satisfy the full TSR condition on only $23.3\%$ ($7/30$) and $0.0\%$ ($0/30$) of targets, respectively. The four general-purpose LLM baselines perform even worse on TSR: Claude and GPT-4o each reach only $6.7\%$ ($2/30$), while o1-mini and o1 both remain at $0.0\%$ ($0/30$), although some still obtain nontrivial DVS \& HQ scores, such as $33.3\%$ ($10/30$) for GPT-4o and o1-mini. These results suggest that isolated returned-set indicators can look acceptable while the full protocol is still not satisfied.

Against this background, CACM substantially improves target-level success over the reproduced DeepSeek-based LIDDiA baseline: TSR increases from $73.3\%$ ($22/30$) to $100.0\%$ ($30/30$). The same trend appears in the returned-set indicators: DVS \& HQ rises from $80.0\%$ ($24/30$) to $100.0\%$ ($30/30$), and $N{\geq}5$ \& DVS rises from $83.3\%$ ($25/30$) to $100.0\%$ ($30/30$). Meanwhile, CACM returns much smaller terminal sets: the average final pool size drops from $21.0$ molecules to exactly $5.0$. This shows that the gain does not come from keeping larger pools until termination, but from returning compact sets that already satisfy the target-level protocol, indicating better stopping quality rather than greater search breadth.

The trajectory view points to the same conclusion. Appendix D reports the cutoff results at 2/4/6/8/10 iterations. The contrast is sharp: the LIDDiA baseline already reaches $90.9\%$ of its eventual successes by iteration~2 ($20/22$), indicating a strongly front-loaded search, whereas CACM gains another $40.0$ percentage points after iteration~2 and reaches full TSR by iteration~8. This suggests that CACM not only improves the endpoint, but also keeps the search productive when early local refinement is insufficient, continuing to extract useful guidance from intermediate failures instead of saturating after the first few actions.

For the LIDDiA baseline and CACM columns, \%m/t and \#m/t denote target-wise averages over the returned pools, and the returned-pool summary medians in the lower part of Table~\ref{tab:main_results} report per-target medians parsed from the logs. Appendix C lists the exact terminal molecule sets for all 30 targets from the same logs used to produce Table~\ref{tab:main_results}. We move these molecule strings out of the main body because the per-target SMILES are too long to present clearly within the space limit.

\paragraph{Returned-pool molecule quality.}
The returned-pool summary medians in the lower part of Table~\ref{tab:main_results} show that several baselines remain competitive on individual molecule-level statistics. DiffSMOL reaches the highest novelty median ($0.89$), Claude gives the highest QED median ($0.78$), several methods reach the top LRF median ($4.00$), and o1-mini obtains the best SAS median ($2.02$). Even the reproduced DeepSeek-based LIDDiA baseline remains reasonably strong on these pooled summaries, with medians of $0.84$ for NVT, $0.69$ for QED, $3.93$ for LRF, $2.67$ for SAS, and $-7.15$ for VNA. CACM improves some of these rows only moderately, for example from $0.69$ to $0.72$ on QED and from $-7.15$ to $-7.17$ on VNA.

This apparent gap between a large TSR improvement and only modest pooled-median gains is expected. These rows summarize individual-molecule properties inside the returned pools, and a baseline can score well here because a few particularly strong molecules pull up one metric or one average, even when the returned set as a whole still fails the TSR protocol. The issue is therefore not whether some molecules are good, but whether the final set jointly satisfies all returned-set conditions at the same time. CACM's main advantage is that it converts comparable molecule-level quality into much stronger set-level reliability, rather than merely maximizing one property in isolation.

\begin{table*}[t]
\centering
\caption{\textbf{Main results on the 30-target LIDDiA benchmark.} Numbers for Pocket2Mol, DiffSMOL, Claude, GPT-4o, o1-mini, and o1 are taken from the original LIDDiA paper; the last two columns are recomputed from our logs.}
\label{tab:main_results}
\footnotesize
\setlength{\tabcolsep}{3.0pt}
\renewcommand{\arraystretch}{1.08}

\resizebox{\textwidth}{!}{%
\begin{tabular}{l|cc|cc|cc|cc|cc|cc|cc|cc}
\toprule
\multirow{2}{*}{Metric}
& \multicolumn{2}{c|}{Pocket2Mol}
& \multicolumn{2}{c|}{DiffSMOL}
& \multicolumn{2}{c|}{Claude}
& \multicolumn{2}{c|}{GPT-4o}
& \multicolumn{2}{c|}{o1-mini}
& \multicolumn{2}{c|}{o1}
& \multicolumn{2}{c|}{LIDDiA (DeepSeek)}
& \multicolumn{2}{c}{CACM (ours)} \\
& \makecell{\%m/t\\or \%t} & \makecell{\#m/t\\or \#t}
& \makecell{\%m/t\\or \%t} & \makecell{\#m/t\\or \#t}
& \makecell{\%m/t\\or \%t} & \makecell{\#m/t\\or \#t}
& \makecell{\%m/t\\or \%t} & \makecell{\#m/t\\or \#t}
& \makecell{\%m/t\\or \%t} & \makecell{\#m/t\\or \#t}
& \makecell{\%m/t\\or \%t} & \makecell{\#m/t\\or \#t}
& \makecell{\%m/t\\or \%t} & \makecell{\#m/t\\or \#t}
& \makecell{\%m/t\\or \%t} & \makecell{\#m/t\\or \#t} \\
\midrule

\multicolumn{17}{l}{\textbf{Generated molecules}} \\
Generated
& -- & 100.0
& -- & 100.0
& -- & 5.0
& -- & 5.0
& -- & 5.0
& -- & 5.0
& -- & 21.0
& -- & 5.0 \\
Valid
& \textbf{100.0} & 100.0
& 99.9 & 99.9
& 98.7 & 4.9
& 97.3 & 4.9
& 91.3 & 4.6
& 95.3 & 4.8
& \textbf{100.0} & 21.0
& \textbf{100.0} & 5.0 \\
QED $\geq \overline{\mathrm{QED}}_t$
& 53.4 & 53.4
& 60.0 & 60.0
& 96.7 & 4.8
& 88.2 & 4.4
& 90.1 & 4.5
& 88.3 & 4.4
& 96.6 & 20.2
& \textbf{100.0} & 5.0 \\
LRF $\geq \overline{\mathrm{LRF}}_t$
& 99.7 & 99.7
& 72.1 & 72.1
& 98.7 & 4.9
& 95.9 & 4.8
& 90.7 & 4.5
& 95.3 & 4.8
& 97.0 & 20.2
& \textbf{100.0} & 5.0 \\
SAS $\leq \overline{\mathrm{SAS}}_t$
& 77.4 & 77.4
& 7.5 & 7.5
& 92.7 & 4.6
& 90.7 & 4.5
& 81.4 & 4.1
& 92.6 & 4.6
& 93.5 & 19.7
& \textbf{100.0} & 5.0 \\
VNA $\leq \overline{\mathrm{VNA}}_t$
& 15.3 & 15.3
& 24.7 & 24.7
& 63.3 & 3.2
& 59.2 & 3.0
& 47.9 & 2.3
& 34.6 & 1.8
& 97.3 & 20.6
& \textbf{100.0} & 5.0 \\
NVT $\geq 0.8$
& 87.6 & 87.6
& 98.2 & 98.2
& 46.9 & 2.4
& 68.3 & 3.4
& 64.1 & 3.2
& 55.9 & 2.8
& 97.0 & 20.3
& \textbf{100.0} & 5.0 \\
HQ
& 6.4 & 6.4
& 0.7 & 0.7
& 30.3 & 1.5
& 35.0 & 1.7
& 28.2 & 1.4
& 20.7 & 1.0
& 88.8 & 18.7
& \textbf{100.0} & 5.0 \\
\midrule

\multicolumn{17}{l}{\textbf{Among all targets}} \\
DVS $> 0.8$
& \textbf{100.0} & \textbf{30}
& \textbf{100.0} & \textbf{30}
& 30.0 & 9
& 90.0 & 27
& 67.7 & 20
& 70.0 & 21
& 86.7 & 26
& \textbf{100.0} & \textbf{30} \\
$N{\geq}5$ \& DVS
& \textbf{100.0} & \textbf{30}
& \textbf{100.0} & \textbf{30}
& 27.7 & 8
& 77.7 & 23
& 43.3 & 13
& 57.7 & 17
& 83.3 & 25
& \textbf{100.0} & \textbf{30} \\
$N{\geq}5$ \& HQ
& 27.7 & 8
& 3.3 & 1
& 23.3 & 7
& 10.0 & 3
& 0.0 & 0
& 3.3 & 1
& 76.7 & 23
& \textbf{100.0} & \textbf{30} \\
DVS \& HQ
& 23.3 & 7
& 10.0 & 3
& 10.0 & 3
& 33.3 & 10
& 33.3 & 10
& 20.0 & 6
& 80.0 & 24
& \textbf{100.0} & \textbf{30} \\
TSR
& 23.3 & 7
& 0.0 & 0
& 6.7 & 2
& 6.7 & 2
& 0.0 & 0
& 0.0 & 0
& 73.3 & 22
& \textbf{100.0} & \textbf{30} \\
\bottomrule
\end{tabular}%
}

\vspace{0.45em}
{\footnotesize\centering\textbf{Returned-pool summary medians}\par}

\vspace{0.20em}
\resizebox{0.88\textwidth}{!}{%
\begin{tabular}{l|cccccccc}
\toprule
Metric
& Pocket2Mol
& DiffSMOL
& Claude
& GPT-4o
& o1-mini
& o1
& LIDDiA (DeepSeek)
& CACM (ours) \\
\midrule
NVT $\uparrow$ & 0.87 & \textbf{0.89} & 0.77 & 0.82 & 0.79 & 0.80 & 0.84 & 0.85 \\
QED $\uparrow$ & 0.51 & 0.55 & \textbf{0.78} & 0.74 & 0.75 & 0.77 & 0.69 & 0.72 \\
LRF $\uparrow$ & \textbf{4.00} & 3.43 & \textbf{4.00} & 3.99 & 3.85 & \textbf{4.00} & 3.93 & \textbf{4.00} \\
SAS $\downarrow$ & 2.46 & 6.15 & 2.30 & 2.16 & \textbf{2.02} & 2.03 & 2.67 & 2.65 \\
VNA $\downarrow$ & -4.74 & -4.23 & -6.69 & -6.56 & -6.31 & -5.97 & -7.15 & \textbf{-7.17} \\
DVS $\uparrow$ & 0.88 & \textbf{0.89} & 0.76 & 0.84 & 0.79 & 0.80 & 0.81 & \textbf{0.89} \\
\bottomrule
\end{tabular}%
}
\end{table*}

\begin{table*}[t]
\centering
\caption{\textbf{Ablation study of CACM.}
TSR is computed under the same LIDDiA success definition for all variants.
DVS \& HQ is computed from the returned pool.
Avg.\ Term.\ Iters is averaged over 30 targets.}
\label{tab:ablation}
\footnotesize
\setlength{\tabcolsep}{5pt}
\renewcommand{\arraystretch}{1.12}
\begin{tabular}{lcccc|cccc}
\toprule
Variant
& \makecell{Repair\\Signal}
& \makecell{Structured\\Memory}
& \makecell{Dynamic\\Compression}
& \makecell{Corrective\\Selection}
& \makecell{TSR\\(\% / \#)}
& \makecell{DVS \& HQ\\(\% / \#)}
& \makecell{Avg.\ Pool\\Size}
& \makecell{Avg.\ Term.\\Iters} \\
\midrule
LIDDiA baseline
& \xmark & \xmark & \xmark & \xmark
& 73.3 / 22
& 80.0 / 24
& 21.0
& 4.40 \\

Set-level Repair Signal
& \cmark & \xmark & \xmark & \xmark
& 86.7 / 26
& 86.7 / 26
& 6.7
& 4.07 \\

\makecell[l]{CACM w/o\\Corrective Selection}
& \cmark & \cmark & \cmark & \xmark
& 83.3 / 25
& 83.3 / 25
& 8.0
& 4.27 \\

\makecell[l]{CACM w/o\\Dynamic Compression}
& \cmark & \cmark & \xmark & \cmark
& 96.7 / 29
& 96.7 / 29
& 8.6
& 3.40 \\

CACM
& \cmark & \cmark & \cmark & \cmark
& \textbf{100.0 / 30}
& \textbf{100.0 / 30}
& \textbf{5.0}
& \textbf{3.07} \\
\bottomrule
\end{tabular}
\end{table*}

\subsection{Ablation Study}
\label{sec:ablation}

Table~\ref{tab:ablation} analyzes which components are responsible for the gain of CACM.
We progressively remove key components in the corrective control loop and examine how returned-set quality changes under the same evaluation protocol.

We first consider the lightest retained variant, \emph{Set-level Repair Signal}, which keeps grounded repair feedback but does not write corrective information back into memory.
This variant improves over the LIDDiA baseline from $73.3\%$ ($22/30$) to $86.7\%$ ($26/30$) TSR, with DVS \& HQ rising from $80.0\%$ ($24/30$) to $86.7\%$ ($26/30$), while the average returned pool size drops from $21.0$ to $6.7$.
This shows that the repair signal is highly effective: making the current failure state \emph{explicitly legible} to the planner already yields a large gain.

We next examine the role of \emph{precision} in corrective write-back.
When corrective memory is constructed but \emph{corrective selection} is removed, TSR drops from $86.7\%$ ($26/30$) to $83.3\%$ ($25/30$), DVS \& HQ decreases by the same amount, and the average returned pool size increases from $6.7$ to $8.0$.
This shows that writing back \emph{more} corrective information is not necessarily better.
Without explicitly selecting the most decision-relevant corrective entries, excessive failure information can become noise and interfere with planning.
This variant still outperforms the LIDDiA baseline, indicating that constructing an explicit corrective channel is itself useful, even if unfiltered write-back is suboptimal.

We then study the contribution of \emph{refinement} in the descriptive state.
Removing dynamic compression keeps target-level success relatively strong, with TSR remaining at $96.7\%$ ($29/30$), but enlarges the returned pool from $5.0$ to $8.6$ molecules.
DVS \& HQ also drops from $100.0\%$ ($30/30$) to $96.7\%$ ($29/30$), while the average termination iterations increase from $3.07$ to $3.40$.
This indicates that dynamic compression is not merely a token-saving device: by distilling raw history into a more compact agent state, it keeps the planner-facing context more focused.

Finally, the full CACM model is the only variant that simultaneously achieves perfect TSR ($100.0\%$, $30/30$), perfect deterministic DVS \& HQ ($100.0\%$, $30/30$), the smallest returned pools ($5.0$), and the fewest termination iterations ($3.07$).
Taken together, the ablations support a coherent picture:
\emph{set-level repair signals} make failures legible, \emph{corrective selection} makes write-back precise, and \emph{dynamic compression} makes the agent state compact and decision-ready.
Only when these components are combined does the corrective loop become both reliable and efficient.

\setlength{\textfloatsep}{6pt plus 1pt minus 2pt}
\begin{figure}[t]
    \centering
    \IfFileExists{baseline_vs_cacm_memory_chars.pdf}{%
        \includegraphics[width=\columnwidth]{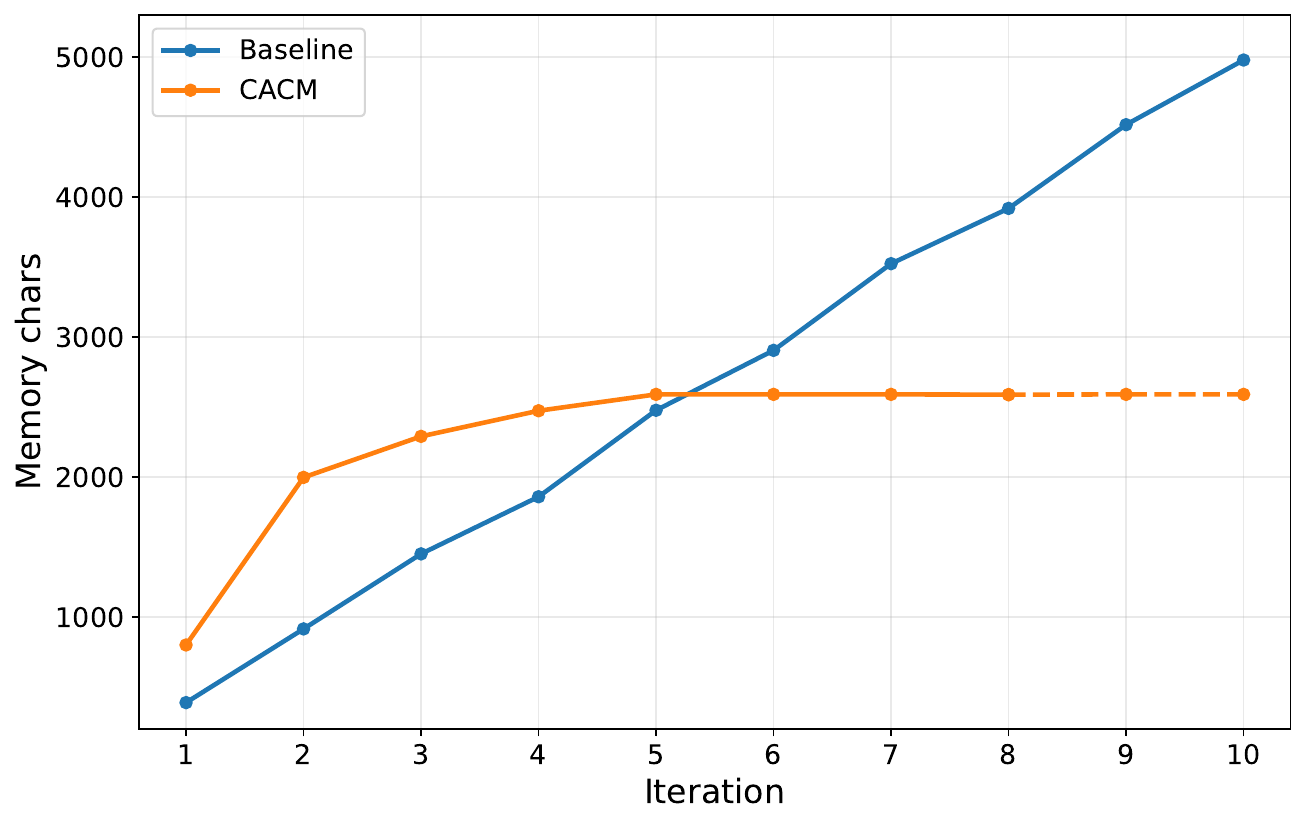}%
    }{%
        \fbox{\parbox{0.98\columnwidth}{\centering
        \textbf{Placeholder for memory compression.}\\[0.4em]
        Plot the average planner-facing memory length over iterations across the 30-target benchmark.\\
        Compare \emph{raw-history write-back} vs.\ \emph{CACM compressed agent state}.}}
    }
    \caption{\textbf{Planner-facing memory length across iterations.}
    Average planner-facing memory length over iterations on the 30-target benchmark, comparing the LIDDiA-style raw-history write-back with the CACM compressed agent state.}
    \label{fig:memory_compression}
\end{figure}

\subsection{Memory Compression}
\label{sec:memory_compression}

Beyond target-level success, we examine how CACM changes the growth of planner-facing agent states.
This matters because CACM does not simply add diagnosis; it rewrites planner context through structured static, dynamic, and corrective channels.
The effect should therefore be assessed not only by final TSR, but also by how the planner-facing agent states evolve over the trajectory.

Figure~\ref{fig:memory_compression} plots the average planner-facing state length over active targets on the 30-target benchmark.
We compare the LIDDiA-style \emph{growing raw-history write-back}, which keeps appending trajectory text over iterations, with the \emph{CACM compressed agent state}, which maintains an organized and compact agent state throughout the search process.
Representative KIT snapshots are shown in Appendix E, while the success-truncated controller-side token and runtime statistics are reported in Appendix F and Appendix G for the corresponding runs in the same evaluation setting.

CACM is not designed to minimize context at the earliest iterations.
As Figure~\ref{fig:memory_compression} shows, it is already longer than the LIDDiA baseline in the first few steps, because it allocates context to protocol-relevant state, diagnosis, and repair guidance.
The key difference is not initial length, but subsequent growth.
The LIDDiA raw-history write-back grows almost monotonically with trajectory length, whereas CACM quickly enters a compact band.
Around iteration~5, the two become comparable in size, after which the LIDDiA baseline continues to expand while CACM remains nearly flat.
The advantage of CACM is thus controlled context growth that preserves task-relevant evidence while preventing planner context from drifting into accumulated raw history over time.

This difference matters for both efficiency and reliability in practice.
A longer free-form history increases prompt cost and exposes the planner to stale or weakly relevant information, whereas CACM preserves a compact agent state centered on protocol-relevant content.
The same pattern appears in end-to-end cost.
Under success-truncated accounting, CACM reduces controller-side tokens per target from 33{,}518.3 to 23{,}042.2, a 31.3\% decrease, and reduces wall-clock time per target from 45.49 to 43.91 minutes, a 3.5\% decrease.
This reduction is enabled by earlier successful stopping: the average cutoff iteration drops from 4.40 to 3.07, a 30.2\% decrease.
At the same time, CACM does not make each step trivially cheaper: controller-side token cost per iteration remains close to the LIDDiA baseline (7{,}945.6 vs.\ 7{,}448.5), while per-iteration runtime is higher because of the extra diagnosis call (15.14 vs.\ 10.11 minutes).
The trajectory-level reduction therefore comes from reaching success in fewer iterations, not from weakening the control process.

This is also consistent with the KIT snapshots in Appendix E.
CACM writes back explicit task requirements, selected pools, recent actions, and corrective guidance as organized planner-facing fields, whereas the LIDDiA baseline exposes a longer appended stream in which old failures, threshold restatements, and local reasoning remain interleaved.
The benefit of CACM is therefore not only shorter memory growth, but also more structured and decision-relevant planner-facing agent states.

This property becomes more important in harder settings.
The current benchmark is short-horizon, so the benefit of structured memory is only partially exposed.
For harder targets, stricter protocols, or longer action horizons, unstructured history would continue to grow with the trajectory, increasing token cost and weakening control quality.
CACM keeps the planner-facing agent state compact, organized, and protocol-focused as the horizon grows.

\subsection{Case Study: How Corrective Memory Changes Control on KIT}
\label{sec:case_kit}

We next examine \textsc{KIT} as a concrete case study of how corrective memory changes control. This target is informative because the system already produces nontrivial candidate pools, yet still fails the returned-set protocol.

Appendix E shows the corresponding planner-facing snapshots. In the CACM snapshot, the retained pools already satisfy the diversity side of the task, with values of $0.859$ and $0.878$, while the worst-case docking value still remains around $-7.358$ against the required threshold of $-7.77$ for this returned pool. The dominant bottleneck is therefore not diversity, but binding: the pool already has enough spread, but still lacks a sufficiently strong binding tail to fully satisfy the set-level criterion at this point.

The contrast with the LIDDiA baseline is structural. The baseline exposes an appended raw-history stream that mixes \textsc{Generate}, \textsc{Code}, and \textsc{Optimize} traces with repeated local judgments, but it does not rewrite the current evidence into an explicit summary of the main protocol gap. CACM instead writes back a selected corrective entry that localizes a \emph{binding bottleneck} and recommends a targeted \textsc{Code}-based repair over the existing pool.

This case clarifies the mechanism of CACM. The gain does not come from writing back more text, but from writing back the right state: task requirements, pocket summary, selected pools, recent actions, and a compressed corrective signal. For \textsc{KIT}, that difference changes planning from open-ended trial-and-error into targeted repair of the dominant bottleneck.

\section{Conclusion}

We presented CACM, a protocol-aware framework for language-based drug discovery agents. Rather than changing the molecular generators, CACM improves how the agent interprets search state, diagnoses returned-set violations, and writes compact corrective signals into memory for planning. By organizing multimodal evidence from task requirements, pocket context, and candidate-set evidence into concise agent states, CACM makes diagnosis more explicit and actions more targeted. On the 30-target LIDDiA benchmark~\cite{averly2025liddia}, CACM improves TSR and the deterministic validity of returned molecule sets, while also reducing trajectory length and final pool size. These results suggest that reliable returned-set optimization remains a missing ingredient in drug discovery agents, and that compact corrective memory provides an effective way to connect auditing, diagnosis, and planning.

\appendix

\section{Metric Definitions and Audit Details}
\label{app:eval_details}

This appendix specifies the metric definitions and deterministic audit procedure used in CACM. Our main text defines each requirement as
\(
r_j=(f_j,\alpha_j,\bowtie_j,b_j)
\),
where $f_j$ is the metric field, $\alpha_j$ is the corresponding set-level aggregation operator, $\bowtie_j$ is the comparison operator, and $b_j$ is the threshold. Given a candidate molecule set
\begin{equation}
S=\{x_i\}_{i=1}^{n},
\end{equation}
the audit computes the observation
\begin{equation}
o_j(S)=\alpha_j(f_j,S)
\end{equation}
for each requirement and then applies the deterministic validity test defined in the main paper.

In CACM, the audit is performed at the \emph{returned-set level}. This means that the final decision is made on the whole returned pool rather than on isolated individual molecules. When a requirement is inherently set-level, the aggregation is applied directly to the pool. When a requirement is defined at the molecule level, we use a worst-case set-level aggregation so that the returned pool is counted as valid only if all required molecules satisfy the corresponding constraint.

\paragraph{Pool-level metrics.}
For pool size, the aggregation is simply the set cardinality,
\begin{equation}
N(S)=|S|.
\end{equation}
For diversity and novelty, we use the same benchmark implementation as the original evaluation pipeline \cite{averly2025liddia} to ensure direct comparability across methods. In the audit, these quantities are treated as set-level observations and compared directly against the required thresholds.

\paragraph{Molecule-level property constraints.}
For lower-bounded molecule-level properties, the audit uses worst-case lower-bound aggregation. For example, if all returned molecules are required to satisfy a QED threshold, the corresponding observation is
\begin{equation}
o_{\mathrm{QED}}(S)=\min_{x\in S}\mathrm{QED}(x),
\end{equation}
so that the returned set passes only when its weakest molecule still satisfies the threshold. QED follows the standard definition of Bickerton \emph{et al.}~\cite{bickerton2012qed}.

For upper-bounded molecule-level properties, the audit uses worst-case upper-bound aggregation. For synthetic accessibility,
\begin{equation}
o_{\mathrm{SAS}}(S)=\max_{x\in S}\mathrm{SAS}(x),
\end{equation}
so that the returned set is counted as valid only when its hardest-to-synthesize molecule still satisfies the bound. Synthetic accessibility is computed using the standard SA score of Ertl and Schuffenhauer~\cite{ertl2009sas}.

For Lipinski-style rule-based drug-likeness constraints, we follow the standard criteria summarized by Lipinski \emph{et al.}~\cite{lipinski2001drug}. When a lower-bound requirement is imposed on the number of satisfied rules or a derived compliance score, the returned-set audit again uses worst-case aggregation over the candidate pool.

For docking, lower scores indicate stronger predicted binding. Therefore, when the protocol requires docking quality to be better than a threshold, the audit uses the worst docking score in the returned pool,
\begin{equation}
o_{\mathrm{Dock}}(S)=\max_{x\in S}\mathrm{Dock}(x),
\end{equation}
and compares it against the required upper bound. Docking is performed using the same AutoDock Vina-based pipeline adopted in our experimental workflow \cite{trott2010autodock}.

\paragraph{Residual-based failure localization.}
Given the observation vector, CACM further computes the residual vector
\(
\Delta^{(k)}
\)
defined in the main paper. These residuals are not used to replace the hard protocol decision; instead, they serve as an explicit failure-localization signal. Negative residuals indicate unmet constraints, and larger magnitudes indicate larger protocol gaps. This residual vector is then used by the diagnosis stage to construct the corrective record used to update corrective memory.

\paragraph{Why deterministic audit matters.}
The purpose of this audit is to ensure that returned-set validity is determined by a transparent and reproducible procedure rather than by free-form language judgment. This is particularly important in our setting because the task objective is defined over the \emph{entire returned set}. A few individually strong molecules are not sufficient if the final pool still violates cardinality, diversity, novelty, docking, or developability constraints. All returned-set results that we recompute from saved trajectories in this paper are evaluated with this unified deterministic audit.

\begin{table}[t]
\centering
\caption{Summary of the main metrics used in the deterministic audit.}
\label{tab:audit_metrics}
\begin{tabular}{l l l}
\toprule
Metric & Aggregation & Ref. \\
\midrule
Pool size & $|S|$ & benchmark\\
Diversity & set-level diversity functional & benchmark \\
Novelty & set-level novelty functional & benchmark \\
QED & $\min_{x\in S}\mathrm{QED}(x)$ & \cite{bickerton2012qed} \\
SAS & $\max_{x\in S}\mathrm{SAS}(x)$ & \cite{ertl2009sas} \\
Lipinski check & worst-case pool aggregation & \cite{lipinski2001drug} \\
Docking & $\max_{x\in S}\mathrm{Dock}(x)$ & \cite{trott2010autodock} \\
\bottomrule
\end{tabular}
\end{table}

\section{Implementation Details of CACM}

This appendix maps the abstract memory operators in Section~3.3 of the main paper to the concrete implementation used in our experiments, and then reports an additional sensitivity study for the implementation-side budgets that realize bounded selection and compression in CACM. Because the main paper describes CACM at the level of memory channels and operators rather than software limits, we first make the implementation-side budget variables explicit here.

At the implementation level, CACM uses two kinds of budgets. \emph{Count budgets} limit how many entries can be preserved before channel-wise formatting. \emph{Character budgets} limit the rendered length of each memory channel after deterministic template formatting. In other words, the implementation introduces bounded memory control at the channel level: static memory is bounded by $B_s$, dynamic descriptive memory is bounded by $B_d$, and corrective memory is bounded by $B_c$. Table~\ref{tab:cacm_budget_symbols} summarizes the symbols used in this section.

\begin{table}[t]
\caption{Implementation-side budget variables used to instantiate bounded selection and compression in CACM.}
\label{tab:cacm_budget_symbols}
\centering
\small
\begin{tabular}{p{0.18\columnwidth}p{0.74\columnwidth}}
\toprule
Symbol & Meaning \\
\midrule
$K_d$ & Maximum number of selected pool summaries retained in dynamic descriptive memory. \\
$W_d$ & Maximum number of retained recent actions in dynamic descriptive memory. \\
$K_c$ & Maximum number of selected corrective entries retained in corrective memory. \\
$B_s$ & Character budget for static memory after deterministic template formatting. \\
$B_d$ & Character budget for dynamic descriptive memory after deterministic template formatting. \\
$B_c$ & Character budget for corrective memory after deterministic template formatting. \\
\bottomrule
\end{tabular}
\end{table}

\paragraph{Why channel-wise budgets are needed.}
The main paper describes CACM as a structured memory mechanism with static, dynamic descriptive, and corrective channels, but it does not spell out how these channels are bounded in software. Without explicit channel-wise budgets, the planner-facing agent state would again drift toward unstructured growth: dynamic descriptive memory could keep accumulating pool summaries and action traces, while corrective memory could keep accumulating past diagnoses even after they are no longer the most decision-relevant signals. The budgets above therefore serve a simple implementation purpose: they turn the abstract selection and compression operators in the main paper into concrete bounded controls for each channel.

\paragraph{Static memory.}
Static memory stores target-specific information that does not change across the trajectory, including the target identity, the normalized requirement set, the pocket file, and a compact pocket summary derived from the receptor structure. Because this channel is trajectory-invariant, its selection step is trivial in implementation. The retained static content is rendered into a fixed template and then bounded by the static-memory budget $B_s$.

\paragraph{Dynamic descriptive memory.}
Dynamic descriptive memory stores the evolving search state. In implementation, this channel is bounded in two stages. First, selection is controlled by a dynamic-pool budget $K_d$ and a recent-action window $W_d$. The former limits how many selected pool summaries are retained; the latter limits how many recent actions are exposed to the planner. Second, the selected dynamic content is rendered into a fixed template and then bounded by the dynamic-memory budget $B_d$.

The selected pool summaries are chosen deterministically rather than by free-form summarization. Concretely, CACM ranks candidate pool summaries using normalized set-level indicators that are already available in the control loop, including pool size, diversity, novelty, QED, Lipinski, docking score, and SAS. The purpose of this ranking is not to redefine the task objective, but to keep the planner-facing dynamic descriptive memory centered on the most decision-relevant pool evidence.

\paragraph{Corrective memory.}
Corrective memory stores failure-localized control information produced by the grounded diagnoser, including the dominant failure pattern, unmet constraints, concise rationale, repair hint, and next-action bias. In implementation, only a bounded number of selected corrective entries are preserved in the planner-facing channel, controlled by the corrective budget $K_c$. These entries are prioritized deterministically using violation severity together with recency, so that the corrective channel emphasizes the most decision-relevant unresolved failures rather than the full history of past diagnoses. The selected corrective content is then rendered into a fixed template and bounded by the corrective-memory budget $B_c$.

\paragraph{Final adaptation.}
After bounded selection and channel-wise formatting, CACM concatenates the labeled static, dynamic descriptive, and corrective channels into a unified planner-facing agent state. No additional global budget variable is introduced at this stage. In practice, compactness is achieved by controlling each channel separately through $B_s$, $B_d$, and $B_c$, together with the entry-selection budgets $K_d$, $W_d$, and $K_c$.

\paragraph{Deterministic formatting.}
The channel-compression functions used in CACM are deterministic template-based formatters followed by truncation under the budgets above. No learned summarizer is introduced at this stage. The implementation-side budgets therefore instantiate the bounded selection and compression operators in the main paper, rather than replacing them with a different learned module.

\begin{table}[t]
\caption{Default implementation-side budgets used in CACM. These values are fixed across the full 30-target benchmark without target-specific tuning.}
\label{tab:cacm_default_budget_values}
\centering
\small
\begin{tabular}{lc}
\toprule
Variable & Value \\
\midrule
$K_d$ & 4 \\
$W_d$ & 3 \\
$K_c$ & 3 \\
$B_s$ & 1400 \\
$B_d$ & 1800 \\
$B_c$ & 1200 \\
\bottomrule
\end{tabular}
\end{table}

\paragraph{Default configuration.}
In the default configuration, CACM uses $K_d=4$, $W_d=3$, and $K_c=3$ for bounded selection, together with $B_s=1400$, $B_d=1800$, and $B_c=1200$ for channel-wise formatting. These values are chosen once at the implementation level and then kept fixed across the full 30-target benchmark, without target-specific tuning.

\paragraph{Sensitivity protocol.}
To test whether the reported CACM performance depends on a narrow implementation-side choice, we ran an additional five-setting sweep around the default configuration. All runs in this section use the same 30-target benchmark, the same controller and molecular toolchain, the same maximum of 10 iterations per target, and the same TSR definition as in the main text; only the implementation-side budgets are changed.

Starting from the default configuration, we construct four nearby variants. Two variants perturb the character budgets. \emph{tight\_chars} compresses all three memory channels. \emph{rebalanced\_chars} reallocates more space to dynamic descriptive memory and less space to static and corrective memory. Two variants perturb the count budgets. \emph{compact\_counts} reduces $(K_d,W_d,K_c)$ from $(4,3,3)$ to $(2,2,2)$, while \emph{wide\_counts} increases them to $(6,5,5)$. This sweep is intended as a local robustness check around the default implementation rather than an exhaustive global hyperparameter search.

\begin{table}[t]
\caption{Budget configurations used in the CACM sensitivity study.}
\label{tab:cacm_budget_settings}
\centering
\small
\begin{tabular}{lcc}
\toprule
Setting & $(K_d, W_d, K_c)$ & $(B_s, B_d, B_c)$ \\
\midrule
default & $(4,3,3)$ & $(1400,1800,1200)$ \\
\emph{tight\_chars} & $(4,3,3)$ & $(900,1200,700)$ \\
\emph{compact\_counts} & $(2,2,2)$ & $(1400,1800,1200)$ \\
\emph{wide\_counts} & $(6,5,5)$ & $(1400,1800,1200)$ \\
\emph{rebalanced\_chars} & $(4,3,3)$ & $(1000,2200,1000)$ \\
\bottomrule
\end{tabular}
\end{table}

\paragraph{Sensitivity results.}
Table~\ref{tab:cacm_budget_results} reports TSR together with average final pool size, average termination iterations, and average planner-facing agent-state length. Table~\ref{tab:cacm_budget_channel_chars} further decomposes the average agent-state length into the static, dynamic descriptive, and corrective channels. Three observations are most relevant.

First, the default configuration is not an isolated success point. In addition to the default setting, \emph{wide\_counts} also reaches $100.0\%$ TSR $(30/30)$, with very similar termination behavior and planner-facing agent-state length. Second, the more aggressive settings remain close to the default rather than collapsing. \emph{tight\_chars}, \emph{compact\_counts}, and \emph{rebalanced\_chars} each still reach $96.7\%$ TSR $(29/30)$. Third, the sweep suggests that the retained-entry budgets are somewhat more consequential than mild character reallocation. In particular, \emph{compact\_counts} is the only setting whose average planner-facing agent-state length becomes slightly longer than the default (1613.4 vs.~1572.2 characters) while also increasing the average final pool size to 8.90. By contrast, \emph{tight\_chars} reduces the average planner-facing agent-state length by 16.2\% (1572.2 $\rightarrow$ 1318.0 characters) yet still preserves $29/30$ target successes.

Taken together, these results support a moderate stability claim: the reported CACM gains do not depend on a narrowly chosen single configuration, and nearby alternatives recover either identical target-level performance or only a one-target drop on the 30-target benchmark. At the same time, the sweep also shows that overly aggressive compression can slightly weaken performance on a small number of harder targets. We therefore interpret the default budgets as reasonably robust implementation-side controls rather than as arbitrary constants with no effect.

\begin{table}[t]
\caption{Sensitivity of CACM to implementation-side budgets on the 30-target benchmark.}
\label{tab:cacm_budget_results}
\centering
\small
\resizebox{\columnwidth}{!}{%
\begin{tabular}{lcccc}
\toprule
Setting & TSR (\% / \#) & Avg. Pool Size & Avg. Term. Iters & Avg. State Chars \\
\midrule
default & 100.0 / 30 & 5.00 & 3.07 & 1572.2 \\
\emph{tight\_chars} & 96.7 / 29 & 8.17 & 3.47 & 1318.0 \\
\emph{compact\_counts} & 96.7 / 29 & 8.90 & 3.50 & 1613.4 \\
\emph{wide\_counts} & 100.0 / 30 & 5.53 & 3.17 & 1580.0 \\
\emph{rebalanced\_chars} & 96.7 / 29 & 8.17 & 3.43 & 1498.4 \\
\bottomrule
\end{tabular}%
}
\end{table}

\begin{table}[t]
\caption{Channel-wise average lengths of the planner-facing agent state in the CACM sensitivity study. Values are averaged over planner-facing states within each trajectory.}
\label{tab:cacm_budget_channel_chars}
\centering
\small
\begin{tabular}{lccc}
\toprule
Setting & Static & Dynamic & Corrective \\
\midrule
default & 601.2 & 241.2 & 729.8 \\
\emph{tight\_chars} & 601.2 & 264.8 & 451.9 \\
\emph{compact\_counts} & 601.2 & 233.0 & 779.3 \\
\emph{wide\_counts} & 601.2 & 256.9 & 721.9 \\
\emph{rebalanced\_chars} & 601.2 & 262.8 & 634.4 \\
\bottomrule
\end{tabular}
\end{table}

\paragraph{Per-target stability.}
The setting-level averages are supported by a simple per-target stability check. Across the 30 benchmark targets, 28 targets remain successful under all five budget settings. Only two targets ever flip across the sweep: \textsc{DRD2} fails only under \emph{compact\_counts}, while \textsc{PTGS1/COX1} fails under \emph{tight\_chars} and \emph{rebalanced\_chars} but succeeds under the other three settings. This means that the observed sensitivity is concentrated on a very small number of harder targets rather than spread broadly across the benchmark. Importantly, these misses are not failed jobs or debugging artifacts: the corresponding runs complete normally and simply reach the iteration cap without achieving a successful stopping point under the same returned-set protocol used in the main text.

\clearpage
\onecolumn

\section{All Returned Molecules for the 30 Targets}
\label{app:all_generated_molecules}

To remove ambiguity about what is actually returned by CACM, this appendix lists the exact terminal molecule set for each of the 30 targets. The molecules reported here are extracted from the same saved trajectories used to compute Table 1 in the main paper, using the same returned-pool selection rule as the main-text audit.

\setlength{\LTpre}{10pt}
\setlength{\LTpost}{8pt}
\small
\renewcommand{\arraystretch}{1.12}

\par\medskip

\begin{longtable}{@{}>{\raggedright\arraybackslash}p{0.13\textwidth}>{\raggedright\arraybackslash}p{0.83\textwidth}@{}}
\caption{Terminal molecules returned by CACM for all 30 targets. Each row corresponds to one target and lists the five molecules in the final returned set.}
\label{tab:all_returned_molecules} \\
\toprule
Target & Returned molecules \\
\midrule
\endfirsthead

\toprule
Target & Returned molecules \\
\midrule
\endhead

\bottomrule
\endfoot

\textbf{ADRB1} & \textbf{1.} \seqsplit{Cn1ccc(-c2cc(C(=O)N3CCN(Cc4ccccn4)CC3)c3ccccc3n2)n1} \par \textbf{2.} \seqsplit{O=C(/C=C/c1ccccc1[N+](=O)[O-])Nc1cccc(Cl)c1} \par \textbf{3.} \seqsplit{Cc1nc(C)c([C@H](C)NC(=O)CCC(=O)c2ccc(Cl)cc2)s1} \par \textbf{4.} \seqsplit{COC(=O)N1CC[C@@H](NC(=O)Nc2cnn(-c3ccccc3F)c2)C1} \par \textbf{5.} \seqsplit{CCn1c(CC[C@@H](C)C=O)nc2ccccc21} \\
\midrule
\textbf{ADRB2} & \textbf{1.} \seqsplit{O=C(CCn1cnc2ccccc2c1=O)OCc1ccc(-c2ccccc2)cc1} \par \textbf{2.} \seqsplit{CC1(c2ccc(OS(=O)(=O)c3ccc(Cl)cc3)cc2)SCCS1} \par \textbf{3.} \seqsplit{Cc1occc1[C@H](Cl)Cc1c(F)cccc1F} \par \textbf{4.} \seqsplit{Cc1ccc(C(=O)N[C@H]2CCC(=O)NC2=O)c(N)c1} \par \textbf{5.} \seqsplit{Cc1cc(C(=O)N2CCN(Cc3ccsc3)CC2)no1} \\
\midrule
\textbf{AR/NR3C4} & \textbf{1.} \seqsplit{Cc1ccc(C[C@H](C)NC(=O)Nc2ccccc2CC(=O)N(C)C)c(C)c1} \par \textbf{2.} \seqsplit{Fc1cccc(F)c1N[C@@H]1CCOC2(CCCC2)C1} \par \textbf{3.} \seqsplit{Cn1c(Cl)cnc1C[S@](=O)CCCc1ccc(Cl)cc1} \par \textbf{4.} \seqsplit{CCc1nc(N2CCOCC2)c2onc(-c3ccc(C)cc3)c2n1} \par \textbf{5.} \seqsplit{O=C(NCCOc1ccc(Br)cc1)[C@H]1CCS(=O)(=O)C1} \\
\midrule
\textbf{BCHE} & \textbf{1.} \seqsplit{O=C(NCC(=O)N1CCOc2ccccc21)NC1CCCCC1} \par \textbf{2.} \seqsplit{Cc1ccc(S(=O)(=O)C[C@H](O)COc2ccc(F)cc2Cl)cc1} \par \textbf{3.} \seqsplit{Cc1sc(NC(=O)c2ccon2)nc1-c1ccc2c(c1)CCC2} \par \textbf{4.} \seqsplit{CNC(=O)c1c(C)cccc1NC(=O)Cc1cccc(OCC\#N)c1} \par \textbf{5.} \seqsplit{O=S(=O)(NCCCn1cncn1)c1cc(Cl)cc(Cl)c1} \\
\midrule
\textbf{CDK5} & \textbf{1.} \seqsplit{COc1ccc(N2C[C@@H](C(=O)N3CCc4ccccc4C3)CC2=O)cc1} \par \textbf{2.} \seqsplit{N\#Cc1ccc(-c2cc(F)cc(C(=O)[O-])c2)cc1O} \par \textbf{3.} \seqsplit{COc1cc(C)c(-c2nc(C3CCCCC3)no2)cc1OC} \par \textbf{4.} \seqsplit{CSc1ccc(Cl)c(C(=O)Nc2cc(C(=O)N(C)C)ccc2Cl)c1} \par \textbf{5.} \seqsplit{C[C@@H](NC(=O)c1cc(=O)c2ccccc2o1)c1cccc(-n2cccn2)c1} \\
\midrule
\textbf{CHK2} & \textbf{1.} \seqsplit{Cc1ccc(NS(=O)(=O)c2cc(C(=O)Nc3ccccc3C(=O)[O-])ccc2Br)cc1} \par \textbf{2.} \seqsplit{c1ccc(SCc2noc(-c3ccco3)n2)nc1} \par \textbf{3.} \seqsplit{O=C(c1ccc(OCC2CC2)nc1)N1CCc2[nH]c(=O)sc2C1} \par \textbf{4.} \seqsplit{COc1cc(Br)c(/C=C2\textbackslash{}SC(=N)NC2=O)cc1OC} \par \textbf{5.} \seqsplit{COc1ccc(CCNC(=O)[C@H]2CC(=O)N(c3cccc(C(F)(F)F)c3)C2)cc1} \\
\midrule
\textbf{CYP3A4} & \textbf{1.} \seqsplit{Cc1ccc([C@@H](C)C(=O)N2CCC[C@H](Cn3nnc4c(O)nc(C)nc43)C2)cc1} \par \textbf{2.} \seqsplit{O=C(N/N=C/C=C\textbackslash{}c1cccc([N+](=O)[O-])c1)C(O)(c1ccccc1)c1ccccc1} \par \textbf{3.} \seqsplit{Clc1cc(I)c(OCc2nc(-c3ccco3)no2)c2ncccc12} \par \textbf{4.} \seqsplit{O=C1C[C@@H](c2ccccc2)[C@@H](c2ccc(Cl)cc2)C(=O)N1} \par \textbf{5.} \seqsplit{COCCNC(=O)c1ccc2c(c1)[C@H]1C=CC[C@H]1[C@@H](c1ccc(O)c(OC)c1)N2} \\
\midrule
\textbf{CYP3A5} & \textbf{1.} \seqsplit{Cc1ccc2cc([C@@H](c3nnnn3C[C@@H]3CCCO3)N3CCN(Cc4ccccc4)CC3)c(=O)[nH]c2c1} \par \textbf{2.} \seqsplit{O=C(Nc1cccc(-c2cnc3ccccc3n2)c1)c1ccc([N+](=O)[O-])cc1} \par \textbf{3.} \seqsplit{O=C1CC2(CCCCC2)[C@@H](c2ccc(F)cc2)C(=O)N1} \par \textbf{4.} \seqsplit{CCc1ccc([C@H](C)NC(=O)NCc2ccc([S@](C)=O)cc2)cc1} \par \textbf{5.} \seqsplit{c1csc(Cn2ccnc2-c2ccnc3ccccc23)c1} \\
\midrule
\textbf{DRD2} & \textbf{1.} \seqsplit{Cc1ccc(N2C[C@H](C(=O)Nc3ccc(C(C)(C)C)cc3)CC2=O)cc1C} \par \textbf{2.} \seqsplit{CC(C)[C@@H](NC(=O)CCc1ccccc1Cl)c1nc2ccccc2[nH]1} \par \textbf{3.} \seqsplit{CC(=O)N(C)c1ccc(NCc2coc3ccccc23)cc1} \par \textbf{4.} \seqsplit{Cc1ccccc1[C@H](CO)Cc1ccc(Cl)c(Cl)c1} \par \textbf{5.} \seqsplit{Cc1cccc(O[C@H](C)CNC(=O)C(=O)Nc2cccnc2Cl)c1} \\
\midrule
\textbf{DRD3} & \textbf{1.} \seqsplit{Cc1csc(C(C)(C)NC(=O)Nc2ccc(F)cc2OCC2CC2)n1} \par \textbf{2.} \seqsplit{CC[C@H](O)c1nc2ccccc2n1CC(C)C} \par \textbf{3.} \seqsplit{Cc1cc(C)cc(-n2ccn(CC(N)=O)c(=O)c2=O)c1} \par \textbf{4.} \seqsplit{Cc1ccc([C@H](NCCc2nnc(-c3ccccc3)o2)C2CC2)cc1} \par \textbf{5.} \seqsplit{O=C(Nc1ccc(-n2cccn2)cc1)c1nc[nH]n1} \\
\midrule
\textbf{EGFR} & \textbf{1.} \seqsplit{O=C([O-])c1cc(Cl)cc(Cl)c1NC(=S)NC(=O)c1cccc2ccccc12} \par \textbf{2.} \seqsplit{CCc1ccc(-c2n[nH]c(SCc3cc(=O)n4ccsc4n3)n2)cc1} \par \textbf{3.} \seqsplit{O=C1C[C@H](NS(=O)(=O)c2ccc3c(c2)OCCO3)CN1c1ccc2c(c1)OCCO2} \par \textbf{4.} \seqsplit{O=C(c1ccc(F)cc1)N1CCC[C@@H](OCc2cc(F)cc(F)c2)C1} \par \textbf{5.} \seqsplit{CC(=O)N1CCc2cc(S(=O)(=O)[C@@H](C)CC(=O)N3CCN(c4ccccc4)CC3)ccc21} \\
\midrule
\textbf{EZH2} & \textbf{1.} \seqsplit{Cc1c2c(cc3c1O/C(=C\textbackslash{}c1c(F)cccc1Cl)C3=O)CN(CCCN1CCOCC1)CO2} \par \textbf{2.} \seqsplit{C[C@@H](Sc1nccn1-c1ccc(Br)cc1)c1nc(N)nc(N(C)C)n1} \par \textbf{3.} \seqsplit{COc1ccc(C)cc1NC(=O)C(=O)N[C@@H](c1ncc(C)s1)C1CC1} \par \textbf{4.} \seqsplit{NS(=O)(=O)c1ccc(NC(=O)CCCSc2ccccc2)cc1} \par \textbf{5.} \seqsplit{CCC[C@@H](C)Nc1nc2cc([N+](=O)[O-])ccc2[nH]1} \\
\midrule
\textbf{FLT3} & \textbf{1.} \seqsplit{Cc1ccc(CCNC(=O)c2csc3c2CC[C@H](C)C3)cc1} \par \textbf{2.} \seqsplit{c1cc(-c2nc(C3CC3)no2)c2cc[nH]c2c1} \par \textbf{3.} \seqsplit{Cc1ccc(-n2cc(-c3cn(Cc4ccoc4)nn3)cn2)cc1C} \par \textbf{4.} \seqsplit{O=C(N[C@H]1CC(=O)N(CC(F)(F)F)C1)[C@H](O)c1ccccc1} \par \textbf{5.} \seqsplit{O=C(NC[C@@H]1COCCO1)c1nnn(-c2ccccc2F)c1-c1ccccn1} \\
\midrule
\textbf{GCK/HK4} & \textbf{1.} \seqsplit{CN(C[C@@H](O)c1ccccc1)C(=O)C1CCN(c2cccc3c2C(=O)N(Cc2ccncc2)C3=O)CC1} \par \textbf{2.} \seqsplit{Cc1ccc(C(=O)NCCNC(=O)Cc2ccc(F)c(F)c2)cc1F} \par \textbf{3.} \seqsplit{CC(C)Cc1nc(CN2CCc3onc(-c4ccc(Cl)cc4)c3C2)no1} \par \textbf{4.} \seqsplit{O=C(Nc1nnc(-c2ccco2)o1)[C@H]1CCCN1S(=O)(=O)c1ccc(F)cc1} \par \textbf{5.} \seqsplit{COc1ccc(OC)c(-c2cc(C(=O)NC[C@]3(O)CCOc4ccccc43)n(C)n2)c1} \\
\midrule
\textbf{GLP1R} & \textbf{1.} \seqsplit{CCN(CC)C(=O)c1cc(C2=CCC(O)(c3ccccn3)CC2)nc2cc(C)ccc12} \par \textbf{2.} \seqsplit{O=c1oc2ccccc2n1C[C@H](c1cccc(C(F)(F)F)c1)C1CC1} \par \textbf{3.} \seqsplit{O=C1C[C@H](C(=O)Nc2ccc(Oc3ccccc3)cc2)NC(N2CCc3ccccc3C2)=N1} \par \textbf{4.} \seqsplit{CC[C@@H](c1nc2c3c(C)c(C)n(-c4ccc(F)cc4)c3ncn2n1)N1C(=O)c2ccccc2C1=O} \par \textbf{5.} \seqsplit{CC1C=CC=C(S(=O)(=O)Nc2ccccc2C(=O)Nc2ccc(F)cc2F)C1} \\
\midrule
\textbf{HRAS} & \textbf{1.} \seqsplit{Cc1ccccc1C(=O)N1CCC(c2nc(O)c3nnn(Cc4ccccc4F)c3n2)CC1} \par \textbf{2.} \seqsplit{CC(C)c1cnn2c(Cl)cc(-c3ccc(Cl)cc3)nc12} \par \textbf{3.} \seqsplit{O=C(/C=C/c1ccc2c(c1)OCO2)Nc1ccc(S(=O)(=O)Nc2nccs2)cc1} \par \textbf{4.} \seqsplit{CC[C@H](NC(=O)CNC(=O)c1cccc(C)c1)C(=O)OC} \par \textbf{5.} \seqsplit{CN(C)c1nc(N)nc(CN(C)S(=O)(=O)c2ccc(F)cc2)n1} \\
\midrule
\textbf{HTR2A} & \textbf{1.} \seqsplit{CCCCn1c([C@H](C)NC(=O)c2ccccc2C)nc2ccccc21} \par \textbf{2.} \seqsplit{N\#Cc1cccc(NC(=O)c2cc3c(s2)CCCC3)c1} \par \textbf{3.} \seqsplit{O=S(=O)(NCCc1ccn(-c2ccccc2)n1)c1ccccc1} \par \textbf{4.} \seqsplit{Cc1cccc(Nc2ccc(S(=O)(=O)N(C(C)C)C(C)C)cn2)c1} \par \textbf{5.} \seqsplit{C=CC(=O)Nc1ccc(-c2nc3cc(NC(=O)C=C)ccc3o2)cc1} \\
\midrule
\textbf{KEAP1} & \textbf{1.} \seqsplit{Cc1csc(-c2ccc(C(=O)Nc3cc(C(C)C)on3)cc2)n1} \par \textbf{2.} \seqsplit{O=S(=O)(c1cccnc1)N1CCC(c2ccc(F)cc2)CC1} \par \textbf{3.} \seqsplit{CC(C)(C)c1ncc2c(n1)CN(c1cc(N3CCOCC3)ncn1)C2} \par \textbf{4.} \seqsplit{CCOc1ccccc1/C=C/C(=O)Nc1ccc(N2CCCCC2)cc1} \par \textbf{5.} \seqsplit{CC1(C)OC(=O)[C@@H](c2ccccc2)O1} \\
\midrule
\textbf{KIT} & \textbf{1.} \seqsplit{Cc1ccc(C2CCC(NCc3cccc(CC(=O)NC4CC4)c3)CC2)cc1} \par \textbf{2.} \seqsplit{Fc1ccc2c(c1)CCN(c1ccc(Cl)cc1)C2} \par \textbf{3.} \seqsplit{Cn1cc(NC(=O)C(=O)c2ccc3nncn3c2)ccc1=O} \par \textbf{4.} \seqsplit{Cc1nc(-c2ccccc2)ccc1-c1ccc(F)cc1} \par \textbf{5.} \seqsplit{O=C1CCc2c(OCc3cccc([N+](=O)[O-])c3)cccc21} \\
\midrule
\textbf{KRAS} & \textbf{1.} \seqsplit{Cc1ccc(S(=O)(=O)N2CCCCC2)cc1NC(=O)[C@@H]1CC(=O)N(c2ccc3c(c2)OCCO3)C1} \par \textbf{2.} \seqsplit{C[C@@H](Cc1cccc(C(F)(F)F)c1)C(=O)N(C)c1ccccn1} \par \textbf{3.} \seqsplit{CC[C@@H]1CCCC[C@H]1OCc1cnc(NN)s1} \par \textbf{4.} \seqsplit{O=C(CN1C(=O)N[C@H](Cc2c[nH]c3ccccc23)C1=O)N1CCC(c2nc3ccccc3s2)CC1} \par \textbf{5.} \seqsplit{COc1cccc(-c2nnc3sc(-c4cc(OC)c(OC)c(OC)c4)nn23)c1} \\
\midrule
\textbf{MET/HGFR} & \textbf{1.} \seqsplit{O=S(=O)(c1ccccc1)c1nnn2c1nc(Nc1cccc(F)c1)c1ccccc12} \par \textbf{2.} \seqsplit{COc1ccc(/C=C2/Cc3c(Cl)cccc3C2=O)cc1C(=O)c1c[nH]c(C)cc1=O} \par \textbf{3.} \seqsplit{C[C@H]1CN(C(=O)[C@@H]2CCN(c3ccccc3)C2)c2ccccc2O1} \par \textbf{4.} \seqsplit{Cc1ccc(-c2nnc(S(=O)(=O)CC(=O)NC3CCCC3)n2C)cc1} \par \textbf{5.} \seqsplit{Cc1ccccc1NC(=O)C(=O)NCCCc1nc2ccccc2[nH]1} \\
\midrule
\textbf{NR3C1/GR} & \textbf{1.} \seqsplit{C[C@H](CNC(=O)[C@H]1C[C@@H]1c1ccccc1)NC(=O)[C@@H]1C[C@@H]1C1=CCC=CC1} \par \textbf{2.} \seqsplit{C=CCN1C(=O)/C(=C/c2ccc(Cl)c(Cl)c2)S/C1=N\textbackslash{}C1C=CC=CN1} \par \textbf{3.} \seqsplit{Cn1cnc(C[NH2+]Cc2ccc(C(C)(C)C)cc2)n1} \par \textbf{4.} \seqsplit{COC(=O)N1CCc2nc(NC(=O)c3cc(Cl)sc3Cl)sc2C1} \par \textbf{5.} \seqsplit{COc1cc([C@@H](N)C(F)(F)F)c(Br)c(Br)c1O} \\
\midrule
\textbf{P2RY12/P2Y12} & \textbf{1.} \seqsplit{C1=Cc2c(ccc3c2CC(c2cc4ccccc4o2)CC3)C1} \par \textbf{2.} \seqsplit{CCn1c(SCC(=O)[O-])nnc1-c1ccc(C)cc1} \par \textbf{3.} \seqsplit{O=c1[nH]c2ccc(NCc3cccc(OCc4ccc(Cl)cc4)c3)cc2[nH]1} \par \textbf{4.} \seqsplit{FC1=CCN(c2nc3nccc(-c4cccc(C(F)(F)F)c4)n3n2)C1} \par \textbf{5.} \seqsplit{C/[NH+]=C(/NCCc1ccc(F)cc1)N1CCC[C@H](C)C1} \\
\midrule
\textbf{PIK3CA} & \textbf{1.} \seqsplit{CCOc1ccc([C@H]2CN(C(=O)Nc3ccc(C)c(C(=O)NC)c3)CCO2)cc1} \par \textbf{2.} \seqsplit{O=C(Cn1ncc(=O)c2ccccc21)NC[C@@H]1CCCN(c2ncc3c(n2)CCCC3)C1} \par \textbf{3.} \seqsplit{Cc1cc(N2CCSCC2)ccc1Nc1cc(-c2cccnc2)nc2ncnn12} \par \textbf{4.} \seqsplit{COc1cccc(CNC(=O)CCc2nnc3ccc(N4C=CCCC4)nn23)c1} \par \textbf{5.} \seqsplit{CC(=O)Nc1c(C)cc(NC(=O)CCc2ccc(F)cc2)cc1C} \\
\midrule
\textbf{PSEN1} & \textbf{1.} \seqsplit{O=C(N[C@@]1(Cc2ccccc2)NC(c2ccccc2)=NC1=O)c1ccccc1} \par \textbf{2.} \seqsplit{COc1ccccc1N1Cc2cccc(N)c2C1} \par \textbf{3.} \seqsplit{Cc1cc(F)c([N+](=O)[O-])cc1S(=O)(=O)NCc1cscn1} \par \textbf{4.} \seqsplit{Cc1noc([C@@H]2CCCN2C(=O)c2cc(C3CC3)on2)n1} \par \textbf{5.} \seqsplit{O[C@H](CN(Cc1ncccc1C(F)(F)F)C1CC1)c1ccccc1} \\
\midrule
\textbf{PTGS1/COX1} & \textbf{1.} \seqsplit{N\#Cc1ccc(C2=CCN(c3ncnc4ccccc34)C2)cc1} \par \textbf{2.} \seqsplit{COc1ccc(F)cc1NC(=O)CCc1ccc2c(c1)OCO2} \par \textbf{3.} \seqsplit{Fc1ccc(N2CC[C@@H](Cc3cccnc3)C2)cc1} \par \textbf{4.} \seqsplit{Cc1cccc(-c2ccc(C\#N)c(C\#N)c2)c1} \par \textbf{5.} \seqsplit{O=C(NCCC(=O)N1CCOCC1)C1CCN(c2ncnc3ccccc23)CC1} \\
\midrule
\textbf{PTGS2/COX2} & \textbf{1.} \seqsplit{Fc1cccc(CCNc2ccccc2)c1} \par \textbf{2.} \seqsplit{Cc1cc(C)c(COc2ccc(CO)nc2)c(C)c1} \par \textbf{3.} \seqsplit{C\#CCNC(=O)CCN1C(=O)COc2ccccc21} \par \textbf{4.} \seqsplit{COc1ccc(/C=C2\textbackslash{}SC(=N)NC2=O)cc1OCc1ccccc1} \par \textbf{5.} \seqsplit{Cc1cc(COc2ccc(CO)nc2)c2cc(Br)ccc2n1} \\
\midrule
\textbf{SHP2/PTPN11} & \textbf{1.} \seqsplit{Cc1ccc(C2=C[C@H](C(=O)N3CCN(C(=O)[C@@H]4NSc5ccc(C)cc54)CC3)N=N2)cc1} \par \textbf{2.} \seqsplit{Clc1nc([C@H]2OCCO[C@@H]2C2=C3CCCC3C3CCCC3=N2)nc2c1CCC2} \par \textbf{3.} \seqsplit{C1=C(COc2ccc(-n3cnnn3)cc2)CC(n2cnnn2)=C1} \par \textbf{4.} \seqsplit{O=S(=O)(NC1=C2CCCC2=CCN1)c1ccccc1F} \par \textbf{5.} \seqsplit{CCC1=CCC(C2=C[C@H](C(=O)c3ccccc3F)N=N2)C=C1} \\
\midrule
\textbf{SLC6A2} & \textbf{1.} \seqsplit{COc1cccc(CC(=O)Nc2ccc3c(c2)C(N)=CCN3C)c1} \par \textbf{2.} \seqsplit{N\#Cc1cccc(C=Cc2cccc(C\#N)c2)c1} \par \textbf{3.} \seqsplit{Cc1cc(N)c(S(=O)(=O)NC2C(C)(C)C2(C)C)c(Cl)c1} \par \textbf{4.} \seqsplit{NC1=C(c2cccc(-c3cccc(N)c3)c2)CC(Cl)=C1} \par \textbf{5.} \seqsplit{Cc1cccc(-c2ncc(NF)cn2)c1} \\
\midrule
\textbf{TNF} & \textbf{1.} \seqsplit{Cc1cc(C(=O)CN2C(=O)N[C@@](C)(c3ccc(Cl)cc3)C2=O)c(C)n1-c1ccccc1} \par \textbf{2.} \seqsplit{COc1cccc(NS(=O)(=O)c2cccc(C(F)(F)F)c2)c1} \par \textbf{3.} \seqsplit{N\#Cc1ccc(OC[C@H](O)CN2CCN(c3ccccc3F)CC2)cc1} \par \textbf{4.} \seqsplit{O=C(Cc1cccc2cccnc12)N(C1CC1)[C@H]1CCS(=O)(=O)C1} \par \textbf{5.} \seqsplit{Cc1nc(S(=O)(=O)N2CCN(Cc3ccc([C@H]4C[C@@H]4C)o3)CC2)cn1C} \\

\end{longtable}

\normalsize
\clearpage
\twocolumn

\section{Additional Main-Result Diagnostics}
\label{app:main_result_diagnostics}

\subsection{Iteration-cutoff comparison at 2/4/6/8/10 iterations}
\label{app:iter_cutoff}

To better understand \emph{when} the performance gain emerges, we report the cutoff results at 2/4/6/8/10 iterations in Table~\ref{tab:ablation_iter_stats}.
We keep this table in the appendix because its main value lies in the trajectory pattern rather than in any single cell.
Consistent with the ablation study in the main text, we retain only the variants used in the final analysis: the LIDDiA baseline, \emph{Set-level Repair Signal}, \emph{CACM w/o corrective selection}, \emph{CACM w/o dynamic compression}, and full CACM.

The table supports the same qualitative story as the main-text ablation.
The LIDDiA baseline is strongly front-loaded: targets that succeed usually succeed early, while targets that fail early seldom recover later.
By contrast, CACM continues to improve after the earliest iterations and exhibits a clear late-stage re-generation pattern.
This is consistent with the claim that the corrective control loop is not merely accelerating the same local refinement behavior, but also helping the agent leave unproductive trajectories and re-enter exploration when the current pool saturates.
The intermediate variants also follow the expected ordering: \emph{Set-level Repair Signal} already improves over the LIDDiA baseline, while removing \emph{corrective selection} or \emph{dynamic compression} weakens either reliability or compactness relative to full CACM.

\begin{table*}[t]
\centering
\scriptsize
\setlength{\tabcolsep}{3pt}
\renewcommand{\arraystretch}{1.05}
\caption{Cutoff statistics at 2/4/6/8/10 iterations for the retained variants in the final ablation.}
\label{tab:ablation_iter_stats}
\resizebox{\textwidth}{!}{%

\begin{tabular}{llccccccccccccc}
\toprule
Method & Iter & Generated & Valid & QED & LRF & SAS & VNA & NVT & HQ & DVS$>0.8$ & $N{\geq}5$\&DVS & $N{\geq}5$\&HQ & DVS\&HQ & TSR \\
\midrule
CACM (full) & 2 & 26.1 & 100.0 / 26.1 & 95.9 / 22.0 & 98.6 / 24.7 & 86.3 / 12.9 & 79.6 / 13.7 & 93.9 / 20.0 & 70.6 / 5.0 & 30/30 & 30/30 & 20/30 & 29/30 & 18/30 \\
 & 4 & 8.2 & 100.0 / 8.2 & 99.8 / 7.9 & 99.8 / 7.9 & 96.3 / 7.0 & 92.9 / 5.5 & 99.1 / 7.3 & 91.5 / 4.7 & 30/30 & 30/30 & 27/30 & 29/30 & 27/30 \\
 & 6 & 11.3 & 100.0 / 11.3 & 98.2 / 9.5 & 99.8 / 11.2 & 98.2 / 9.6 & 95.6 / 7.0 & 96.8 / 8.2 & 93.7 / 5.0 & 30/30 & 30/30 & 29/30 & 29/30 & 28/30 \\
 & 8 & 5.0 & 100.0 / 5.0 & 100.0 / 5.0 & 100.0 / 5.0 & 100.0 / 5.0 & 100.0 / 5.0 & 100.0 / 5.0 & 100.0 / 5.0 & 30/30 & 30/30 & 30/30 & 30/30 & 30/30 \\
 & 10 & 5.0 & 100.0 / 5.0 & 100.0 / 5.0 & 100.0 / 5.0 & 100.0 / 5.0 & 100.0 / 5.0 & 100.0 / 5.0 & 100.0 / 5.0 & 30/30 & 30/30 & 30/30 & 30/30 & 30/30 \\
\midrule
\makecell[l]{CACM w/o\\dynamic compression} & 2 & 18.4 & 100.0 / 18.4 & 94.8 / 13.3 & 99.0 / 17.5 & 95.0 / 13.4 & 88.3 / 10.6 & 95.2 / 13.7 & 83.3 / 5.6 & 30/30 & 30/30 & 27/30 & 29/30 & 24/30 \\
 & 4 & 5.7 & 100.0 / 5.7 & 100.0 / 5.7 & 100.0 / 5.7 & 94.7 / 5.4 & 95.3 / 5.4 & 100.0 / 5.7 & 94.0 / 5.4 & 30/30 & 29/30 & 27/30 & 28/30 & 27/30 \\
 & 6 & 12.1 & 100.0 / 12.1 & 96.8 / 8.9 & 99.7 / 11.8 & 95.8 / 7.9 & 97.0 / 9.1 & 96.1 / 8.2 & 93.4 / 5.5 & 30/30 & 30/30 & 28/30 & 29/30 & 28/30 \\
 & 8 & 5.7 & 100.0 / 5.7 & 100.0 / 5.7 & 100.0 / 5.7 & 100.0 / 5.7 & 100.0 / 5.7 & 100.0 / 5.7 & 100.0 / 5.7 & 30/30 & 29/30 & 29/30 & 30/30 & 29/30 \\
 & 10 & 5.7 & 100.0 / 5.7 & 100.0 / 5.7 & 100.0 / 5.7 & 100.0 / 5.7 & 100.0 / 5.7 & 100.0 / 5.7 & 100.0 / 5.7 & 30/30 & 29/30 & 29/30 & 30/30 & 29/30 \\
\midrule
\makecell[l]{CACM w/o\\corrective selection} & 2 & 14.2 & 100.0 / 14.2 & 91.9 / 11.5 & 97.8 / 13.6 & 83.1 / 8.4 & 79.5 / 8.8 & 94.2 / 12.8 & 65.8 / 4.2 & 29/30 & 29/30 & 17/30 & 24/30 & 13/30 \\
 & 4 & 11.9 & 100.0 / 11.9 & 95.7 / 10.4 & 98.4 / 11.5 & 88.9 / 7.8 & 85.3 / 8.8 & 94.1 / 9.2 & 75.7 / 4.7 & 29/30 & 28/30 & 22/30 & 26/30 & 20/30 \\
 & 6 & 6.0 & 100.0 / 6.0 & 97.5 / 5.5 & 100.0 / 6.0 & 95.0 / 5.0 & 97.5 / 5.5 & 97.0 / 5.4 & 93.3 / 4.6 & 28/30 & 26/30 & 26/30 & 26/30 & 24/30 \\
 & 8 & 6.4 & 100.0 / 6.4 & 96.7 / 5.8 & 99.5 / 6.3 & 95.5 / 5.5 & 97.0 / 5.8 & 95.0 / 5.4 & 90.8 / 4.6 & 29/30 & 28/30 & 27/30 & 27/30 & 25/30 \\
 & 10 & 8.0 & 100.0 / 8.0 & 97.3 / 7.0 & 99.5 / 7.9 & 94.6 / 5.9 & 95.1 / 6.2 & 97.5 / 6.8 & 90.8 / 4.6 & 28/30 & 27/30 & 27/30 & 26/30 & 25/30 \\
\midrule
\makecell[l]{Set-level Repair\\Signal} & 2 & 10.7 & 96.7 / 10.7 & 93.2 / 9.9 & 95.6 / 10.3 & 87.4 / 8.1 & 80.9 / 6.4 & 92.0 / 9.7 & 73.2 / 4.7 & 28/30 & 28/30 & 19/30 & 25/30 & 18/30 \\
 & 4 & 7.6 & 100.0 / 7.6 & 94.0 / 6.4 & 97.0 / 7.0 & 92.8 / 6.2 & 93.8 / 6.5 & 95.5 / 6.7 & 87.2 / 5.2 & 28/30 & 26/30 & 23/30 & 27/30 & 23/30 \\
 & 6 & 6.6 & 100.0 / 6.6 & 97.8 / 6.1 & 98.8 / 6.3 & 93.5 / 5.7 & 98.0 / 6.4 & 97.5 / 6.1 & 89.8 / 5.1 & 28/30 & 26/30 & 23/30 & 27/30 & 23/30 \\
 & 8 & 8.9 & 100.0 / 8.9 & 97.1 / 7.7 & 98.8 / 8.6 & 94.7 / 7.2 & 91.7 / 7.3 & 95.7 / 7.6 & 85.4 / 5.3 & 29/30 & 28/30 & 24/30 & 27/30 & 24/30 \\
 & 10 & 6.7 & 100.0 / 6.7 & 98.0 / 6.3 & 99.7 / 6.7 & 97.5 / 6.2 & 98.5 / 6.4 & 97.3 / 6.2 & 94.7 / 5.7 & 27/30 & 27/30 & 28/30 & 27/30 & 26/30 \\
\midrule
LIDDiA (DeepSeek) & 2 & 21.3 & 96.7 / 21.3 & 90.2 / 17.5 & 90.2 / 17.5 & 90.2 / 17.5 & 90.2 / 17.5 & 90.2 / 17.5 & 90.2 / 17.5 & 27/30 & 23/30 & 22/30 & 26/30 & 20/30 \\
 & 4 & 22.6 & 100.0 / 22.6 & 93.5 / 18.8 & 93.5 / 18.8 & 93.5 / 18.8 & 93.5 / 18.8 & 93.5 / 18.8 & 93.5 / 18.8 & 28/30 & 25/30 & 25/30 & 26/30 & 22/30 \\
 & 6 & 19.4 & 100.0 / 19.4 & 98.8 / 19.1 & 99.2 / 19.2 & 98.3 / 19.0 & 96.7 / 18.7 & 98.0 / 19.0 & 94.7 / 18.3 & 24/30 & 22/30 & 21/30 & 22/30 & 20/30 \\
 & 8 & 19.7 & 100.0 / 19.7 & 98.7 / 19.5 & 99.8 / 19.7 & 97.0 / 19.1 & 97.8 / 19.3 & 96.3 / 19.0 & 93.7 / 18.5 & 25/30 & 24/30 & 22/30 & 23/30 & 21/30 \\
 & 10 & 21.0 & 100.0 / 21.0 & 96.6 / 20.2 & 97.0 / 20.2 & 93.5 / 19.7 & 97.3 / 20.6 & 97.0 / 20.3 & 88.8 / 18.7 & 26/30 & 25/30 & 23/30 & 24/30 & 22/30 \\
\bottomrule
\end{tabular}%
}
\end{table*}

\begin{table*}[t]
\centering
\caption{\textbf{Success-truncated controller-side token usage on the 30-target benchmark.}
For each target, tokens are accumulated only up to the first successful stopping point; if no success is reached, the full 10-iteration budget is used.
``Extra'' denotes the metric-judgment stage for the LIDDiA baseline and the diagnosis stage for CACM.}
\label{tab:token_overhead_success}
\small
\setlength{\tabcolsep}{5pt}
\renewcommand{\arraystretch}{1.08}
\begin{tabular}{lccccccc}
\toprule
Method
& \makecell{Planner\\ / target}
& \makecell{Extra\\ / target}
& \makecell{Extra\\ share}
& \makecell{Total\\ / target}
& \makecell{Planner\\ / iter}
& \makecell{Extra\\ / iter}
& \makecell{Total\\ / iter} \\
\midrule
LIDDiA baseline
& 24,799.0
& 8,719.3
& 26.0\%
& 33,518.3
& 5,510.9
& 1,937.6
& 7,448.5 \\
CACM
& 9,927.9
& 13,114.4
& 56.9\%
& 23,042.2
& 3,423.4
& 4,522.2
& 7,945.6 \\
\bottomrule
\end{tabular}
\end{table*}

\begin{table}[t]
\centering
\caption{\textbf{Success-truncated wall-clock runtime on the 30-target benchmark.}
For each target, runtime is accumulated only up to the first successful stopping point; if no success is reached, the full 10-iteration budget is used.}
\label{tab:runtime_overhead_success}
\footnotesize
\setlength{\tabcolsep}{3.5pt}
\renewcommand{\arraystretch}{1.0}
\begin{tabular}{lccc}
\toprule
Method
& \makecell{Avg.\ cutoff\\ iterations}
& \makecell{Minutes\\ / iter}
& \makecell{Minutes\\ / target} \\
\midrule
LIDDiA baseline & 4.40 & 10.11 & 45.49 \\
CACM            & 3.07 & 15.14 & 43.91 \\
\bottomrule
\end{tabular}
\end{table}

\section{Representative Planner-Facing Memory on KIT}
\label{app:memory_vis}

This appendix juxtaposes representative planner-facing memory snapshots for the target KIT under CACM and the LIDDiA baseline.
We choose this case because it makes the control difference directly visible.
Both methods expose substantial trajectory information to the planner, but they organize that information very differently.

The CACM snapshot shows a compact and decision-relevant control state that explicitly summarizes task constraints, pocket context, selected molecule pools, recent actions, and corrective guidance derived from the current failure mode.
The LIDDiA baseline snapshot, by contrast, shows the same target through a chronologically accumulated planner-facing memory stream that interleaves task restatement, planner reasoning, chosen actions, action outputs, evaluation summaries, and success judgments.
Importantly, the baseline snapshot is shown at iteration~6/10 as an intermediate failure-case state rather than a successful stopping point.

The purpose of presenting both snapshots is not simply to compare memory length, but to make visible the difference between \emph{information accumulation} and \emph{control-oriented write-back}.

\paragraph{What is visible in the planner-facing memory.}
Readers can directly verify that both snapshots contain nontrivial planner-visible evidence in plain text.
The key difference is organizational.
CACM writes back explicit control fields such as \emph{Task requirements}, \emph{Pocket summary}, \emph{Current selected molecule pools}, \emph{Recent actions}, and \emph{Selected corrective entries}.
The LIDDiA baseline instead exposes a single appended planner-facing memory stream in which task restatement, action traces, local judgments, and success checks remain interleaved.
This contrast is important because it shows that CACM does not merely shorten memory; it rewrites available evidence into a more decision-relevant planner state.

\subsection{CACM snapshot: KIT at iteration 4}

\begin{quote}
\small
\noindent\textbf{Target:} KIT \\
\textbf{Method:} CACM \\
\textbf{Iteration shown:} 4 (raw step 3) \\
\textbf{Total memory length:} 2,389 characters \\
\textbf{Channel lengths:} static 598; dynamic 596; corrective 1,195.

\medskip
\noindent\textbf{Task requirements} \\
At least 5 molecules. \\
Vina score must be lower than $-7.77$. \\
Novelty must be at least $0.80$. \\
Diversity must be at least $0.80$. \\
QED must be better than $0.43$. \\
SAScore must be better than $2.77$. \\
Lipinski must be better than or at least $3.19$.

\medskip
\noindent\textbf{Pocket summary} \\
Atom count: 277; residue count: 36; chain count: 1. \\
Bounding-box size: $[20.348,\ 24.401,\ 24.296]$. \\
Pocket center: $[34.162,\ 14.721,\ 39.277]$. \\
Top residue types: LEU(5), VAL(5), THR(4), ILE(4), LYS(3). \\
Hydrophobic ratio: 0.528; positive ratio: 0.111; negative ratio: 0.083; aromatic ratio: 0.111.

\medskip
\noindent\textbf{Current selected molecule pools} \\
\textbf{MOL003}: score 0.6413; size 100; diversity 0.8487; worst Vina $-3.676$; minimum novelty 0.685; minimum QED 0.306; maximum SAS 2.769. \\
\textbf{MOL001}: score 0.6615; size 100; diversity 0.8781; worst Vina $-3.932$; minimum novelty 0.695; minimum QED 0.343; maximum SAS 5.229. \\
\textbf{MOL002}: score 0.6570; size 100; diversity 0.8592; worst Vina $-7.358$; minimum novelty 0.685; minimum QED 0.212; maximum SAS 4.682.

\medskip
\noindent\textbf{Recent actions} \\
Iteration-history entry 0: \textsc{Generate} $\rightarrow$ MOL001; strict protocol pass = False. \\
Iteration-history entry 1: \textsc{Optimize} $\rightarrow$ MOL002; strict protocol pass = False. \\
Iteration-history entry 2: \textsc{Optimize} $\rightarrow$ MOL003; strict protocol pass = False.

\medskip
\noindent\textbf{Selected corrective entry} \\
Iteration 2; failure family: binding bottleneck; severity: 4.094; recommended bias: \textsc{Code}. \\
Failed requirements: Vina score $< -7.77$, novelty $\ge 0.80$, QED $> 0.43$, and Lipinski $\ge 3.19$. \\
Repair hint: the current pool already contains 24 molecules that satisfy all per-molecule thresholds, so the preferred next step is a code-based repair over the existing pool: filter molecules by Vina score, novelty, QED, SAScore, and Lipinski constraints, then construct a diverse subset with at least five molecules.

\medskip
\noindent\emph{Note.} The raw snapshot retains two identical corrective entries at this point in the trajectory. For readability, we show the duplicated entry only once here.
\end{quote}

\paragraph{Why this CACM example matters.}
This reformatted snapshot makes two points concrete.
First, CACM does not simply record that previous attempts failed; it exposes \emph{which} constraints remain unsatisfied and turns them into an explicit repair instruction.
Second, the write-back remains structured and compact: instead of feeding the planner a long unfiltered trajectory, CACM surfaces the most decision-relevant pools, actions, and corrective signal in a form that can directly guide the next step.

\subsection{LIDDiA baseline snapshot: KIT at iteration 6/10}
\label{app:baseline_memory_vis}

\begin{quote}
\small
\noindent\textbf{Target:} KIT \\
\textbf{Method:} LIDDiA baseline \\
\textbf{Iteration shown:} 6/10 (failure-case snapshot) \\
\textbf{Approximate displayed memory length:} 2,904 characters \\
\textbf{Organization:} single appended planner-facing memory stream; no static/dynamic/corrective channel split.

\medskip
\noindent\textbf{Planner-facing memory (faithfully reformatted for display, original chronological style preserved)}

\medskip
\noindent
Initial task: design a set of molecules for KIT. Requirements: at least 5 molecules; Vina score lower than $-7.77$; novelty at least $0.80$; diversity at least $0.80$; QED better than $0.43$; SAScore better than $2.77$; Lipinski better than or at least $3.19$. Available actions: \textsc{Generate}, \textsc{Optimize}, \textsc{Code}.

\medskip
\noindent
Iteration 1. Planner response: start with \textsc{Generate} because no candidate pool is available yet. Action: \textsc{Generate} $\rightarrow$ MOL001. Evaluation after iteration 1: requirements not satisfied. Failed constraints include Vina score, novelty, QED, SAScore, and Lipinski. Current conclusion: generated molecules are diverse enough to explore, but binding and several per-molecule quality indicators remain weak.

\medskip
\noindent
Iteration 2. Planner response: since binding and quality remain poor, try filtering the current pool by the known thresholds and keep the better subset. Action: \textsc{Code} on MOL001 $\rightarrow$ MOL002. Evaluation after iteration 2: only 3 molecules remain after filtering, so the pool now fails the size requirement even though some molecule-level properties improve. Current conclusion: threshold-based filtering may help quality, but it can also shrink the pool too aggressively.

\medskip
\noindent
Iteration 3. Planner response: because direct filtering left too few molecules, try improving the pool itself, especially docking, before filtering again. Action: \textsc{Optimize} on MOL001 with Vina-oriented objective $\rightarrow$ MOL003. Evaluation after iteration 3: requirements still not satisfied. Failed constraints again include Vina score, novelty, QED, SAScore, and Lipinski. Current conclusion: optimization changes the pool, but the dominant bottleneck is still unclear in the appended memory because multiple failed constraints keep appearing together.

\medskip
\noindent
Iteration 4. Planner response: combine information from MOL001, MOL002, and MOL003, then filter using the target constraints again. Action: \textsc{Code} on MOL001, MOL002, and MOL003 $\rightarrow$ MOL004. Evaluation after iteration 4: deterministic metric check reports that the task requirements are satisfied. Success judgment text appended to memory: \emph{Answer: NO}. Reason given in the memory: QED is around $0.43$, and the planner is not fully confident that the strict inequality ``better than $0.43$'' is satisfied for the returned pool.

\medskip
\noindent
Iteration 5. Planner response: the main issue may now be the QED boundary rather than binding, so apply another code-based filter to keep only molecules with QED strictly above $0.43$, while still trying to preserve the set size and diversity conditions. Action: \textsc{Code} on MOL004 $\rightarrow$ MOL005. Evaluation after iteration 5: deterministic metric check again reports that the task requirements are satisfied. Success judgment text appended to memory: \emph{Answer: NO}. Reason again focuses on the strict QED boundary and confidence about whether every returned molecule exceeds the threshold.

\medskip
\noindent
Iteration 6. The planner-facing memory at this point already contains the full original task specification, repeated mentions of the same thresholds, two code-filtering attempts, one optimization attempt, repeated statements that several constraints fail together, and then repeated local reasoning about whether QED is strictly greater than $0.43$. The latest planner response therefore mixes several concerns at once: maybe continue coding because the QED boundary is still uncertain; maybe optimize again because binding was poor earlier; maybe combine multiple pools because filtering alone may reduce the set size. The memory has become longer and more detailed, but the dominant failure mode is still not written back as an explicit corrective summary.

\medskip
\noindent
Visible characteristics of the appended memory at this point: old failures and recent local issues coexist in the same stream; requirement text is restated multiple times; the planner sees action logs and evaluation summaries, but no ranked retained pools, no compact recent-action abstraction, and no selected corrective entry.
\end{quote}

\paragraph{Why this baseline example matters.}
This snapshot clarifies the contrast with the CACM memory above.
The issue is not that the LIDDiA baseline lacks memory content; on the contrary, even at this intermediate failure-case state, the planner already sees a substantial amount of accumulated text.
The issue is that the write-back remains an appended trajectory memory rather than a compact control state.
As a result, evidence about the current bottleneck is present, but it is mixed with repeated thresholds, past local failures, and free-form intermediate reasoning.
CACM, by contrast, converts such evidence into a bounded and explicitly decision-oriented planner-visible state.

\section{Success-Truncated Controller-side Token Usage}
\label{app:token_overhead}

We additionally analyze controller-side token usage under a success-truncated protocol.
For each target, we accumulate all logged controller-side tokens only up to the first successful stopping point recorded in the trajectory; if no success is reached, we count the full 10-iteration budget.
This yields a fairer measure of practical cost than averaging over forced full-length runs.

Table~\ref{tab:token_overhead_success} separates the controller-side budget into the \emph{planner} and an \emph{extra controller-side stage}.
For the LIDDiA baseline, this extra cost comes from the metric-judgment call used to assess whether the current pool satisfies the stopping condition.
For CACM, it comes from the additional diagnosis call.
The table reports both the average token usage per target until success and the corresponding average per active iteration.

Two observations are important.
First, the LIDDiA baseline is not a single-call controller in practice: its metric-judgment stage already contributes a nontrivial fraction of total controller-side cost, accounting for $26.0\%$ of the token budget up to success.
Second, CACM does introduce a larger auxiliary controller-side share, but at the same time sharply reduces planner consumption.
Consequently, although the two methods are close on a per-iteration basis, CACM uses fewer controller-side tokens overall when measured up to success at the target level.

\section{Success-Truncated Runtime Overhead}
\label{app:runtime_overhead}

We next report wall-clock runtime under the same success-truncated protocol.
For each target, runtime is measured only up to the first successful stopping point; if the target does not succeed, we count the full 10-iteration runtime.
We then average both the target-level runtime and the per-iteration runtime over the 30-target benchmark.

Table~\ref{tab:runtime_overhead_success} shows the same trade-off pattern as the token analysis.
CACM is slower for each active iteration because it adds an extra controller-side call.
However, CACM reaches successful stopping in fewer iterations on average, which lowers the overall time spent per target until success.
This supports the claim that the bounded corrective controller is not merely paying extra time at every step without return; rather, it trades a moderate per-iteration overhead for lower end-to-end cost at the trajectory level.

%
%
% ...



\clearpage

\clearpage
\bibliographystyle{ACM-Reference-Format}
\bibliography{references}
\clearpage
\end{document}